\documentclass[]{rtp_tech_report}

\usepackage{wrapfig}
\usepackage{float}
\tcbuselibrary{breakable,skins,listings}

\newtcblisting{templatebubble}{
  enhanced,
  breakable,
  listing only,
  colback=orange!4,
  colframe=orange!75!black,
  boxrule=0.5pt,
  arc=7pt,
  left=7pt,
  right=7pt,
  top=6pt,
  bottom=6pt,
  before skip=0.85em,
  after skip=1.0em,
  listing options={
    basicstyle=\ttfamily\footnotesize,
    breaklines=true,
    breakindent=0pt,
    breakautoindent=false,
    numbers=left,
    numberstyle=\scriptsize\color{orange!70!black},
    numbersep=7pt,
    xleftmargin=2.1em,
    columns=fullflexible,
    keepspaces=true,
    showstringspaces=false,
    showlines=true
  }
}

\title{Text Template Tokens Are Implicit Semantic Registers in Diffusion Transformers}

\author[1,2,\sharp]{Maohua Li}
\author[2,3,\sharp]{Qirui Li}
\author[1]{Yanke Zhou}
\author[2]{Yiduo Li}
\author[2]{Zhaosheng Chi}
\author[2]{Chao Xu}
\author[2]{Cuifeng Shen}
\author[2]{Yixuan Xu}
\author[2,\dagger]{Hanlin Tang}
\author[2]{Kan Liu}
\author[2]{Tao Lan}
\author[2]{Lin Qu}
\author[1,\S]{Shao-Qun Zhang}

\affiliation[1]{Nanjing University}
\affiliation[2]{Alibaba Group}
\affiliation[3]{Zhejiang University}

\contribution[\S]{Corresponding author}
\contribution[\dagger]{Project lead}
\contribution[\sharp]{Work done during internship at Alibaba}
\checkdata[E-mail]{\texttt{\{limaohua, zhangsq\}@lamda.nju.edu.cn}}
\checkdata[Code]{\url{https://github.com/Met4physics/DiT-Interpretability}}

\abstract{
Modern text-to-image diffusion transformers (DiTs) generate images through
joint attention, in which text and image tokens interact directly within a
single sequence. In large-scale DiTs, the conditioning input contains not only
the user prompt but also chat-template tokens introduced by LLM-based text
encoders. Yet how these tokens participate in the denoising computation remains
poorly understood. To probe this, we introduce a causal interpretability
framework. Using it to separate prompt-content tokens from chat-template tokens,
we find that the template tokens carry little prompt-specific
information at the encoder output. Yet surprisingly, they emerge as dominant
image-to-text attention sinks and causally maintain object identity inside the
DiT, acting as implicit semantic registers. We show that they acquire this
identity indirectly. Rather than reading the prompt tokens, they draw the
identity from the image latents into which the prompt semantics have already
been injected at the very first layer. We further reveal a division of labor
across heads and depth in DiTs, where distinct heads route semantics or render
visual structure, and identity is committed in early blocks, carried by middle
blocks, and refined in late ones.
As a practical payoff, this analysis yields a training-free pruning rule that
removes the causally inert prompt-reading heads and cuts $20\%$ of
joint-attention FLOPs at a $1.4$-point cost in GenEval accuracy.
Overall, our work not only reveals that the tokens encoding semantics at the input need not be those that maintain them during generation, but also provides a causal view of internal mechanisms in diffusion transformers.
}

\begin{document}

\maketitle

\section{Introduction}

Modern text-to-image (T2I) models exhibit remarkable text-following
ability~\citep{esser2024scaling,flux2024,wu2025qwen}.
They can preserve object identity, attributes, relations, and compositional
constraints while transforming noise into a coherent image. This capability
raises a fundamental mechanistic question:
\begin{insightline}
How is information from the text actually incorporated into a
diffusion transformer (DiT)?
\end{insightline}
\noindent The natural picture is one of direct and repeated injection. At every layer, image tokens
attend to the prompt tokens that explicitly encode the requested content, so
the semantics remain anchored to their textual source throughout denoising.
Despite its intuitive appeal, this picture has not been causally established.
We show that it is also incomplete: semantic incorporation in DiTs is far more
indirect and structured than layer-wise prompt reading.

\begin{figure}[H]
\centering
\includegraphics[width=\textwidth]{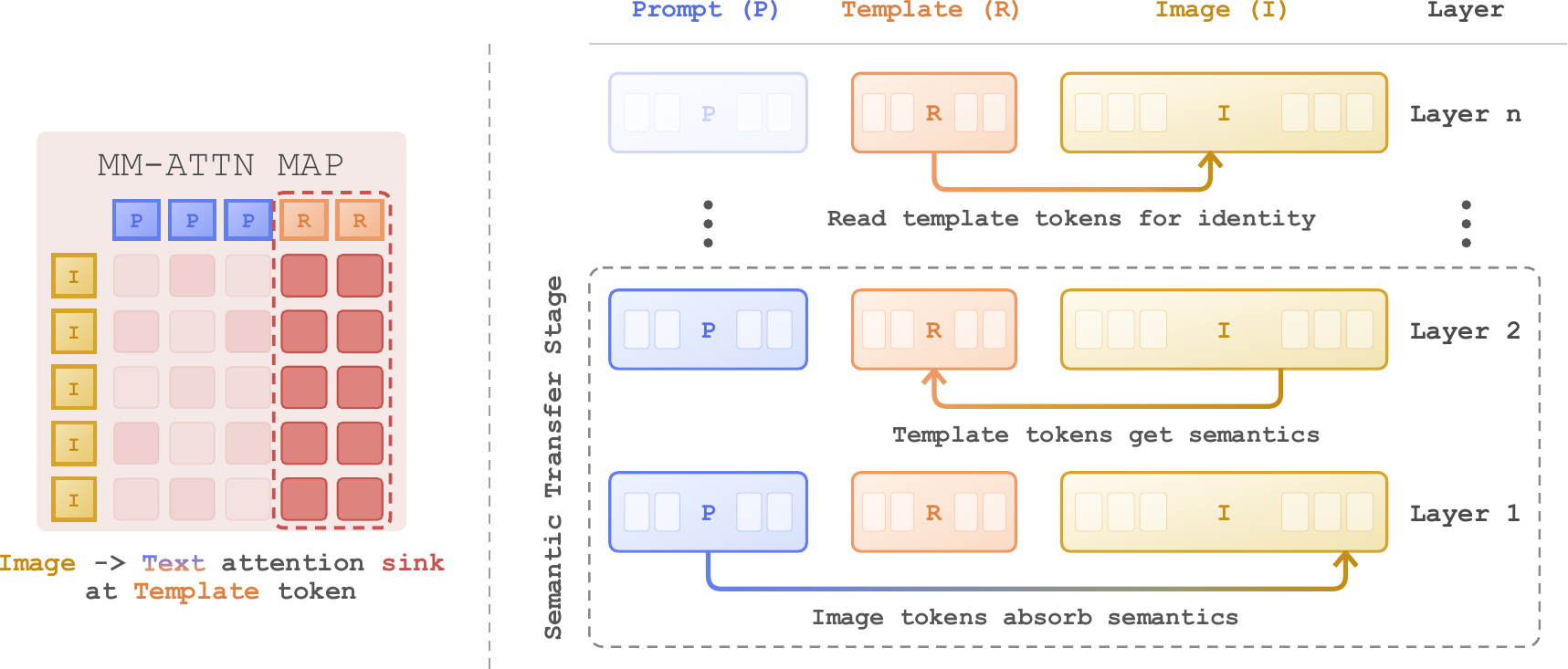}
\caption{Our core findings. \textbf{Left:} in the
multimodal attention map, the attention that image tokens send to
the text stream concentrates on the chat-template tokens
rather than on the semantic prompt tokens, an
attention-sink pattern. \textbf{Right:} the semantics transfer mechanism.
In the first layer image tokens absorb semantics directly from the prompt tokens,
and in the second layer template tokens read object identity from the image span.
}
\label{fig:teaser}
\end{figure}

State-of-the-art T2I systems increasingly build on DiT
backbones~\citep{peebles2023scalable} with joint attention, where text and image
tokens interact directly~\citep{esser2024scaling}. The most recent of these
systems further replace CLIP/T5 text encoders with LLM- or VLM-based encoders
that wrap prompts in chat templates~\citep{wu2025qwen,flux2bfl}. Their
conditioning therefore contains both semantic prompt tokens $\mathcal{S}$ and
structural chat-template tokens $\mathcal{R}$, which we call \emph{template
tokens}. Because every conditioning token can interact with the image tokens
$\mathcal{I}$, it remains unclear which tokens actually incorporate and
maintain semantics during generation. In this work, we systematically study this question, asking whether
deeper DiT layers continue to read the prompt directly or instead rely on other
internal carriers of semantic information.

Our central finding overturns the direct-injection picture. Although template
tokens contain little prompt-specific information at the text-encoder output,
they develop a striking attention-sink pattern~\citep{xiao2024efficient} inside the DiT: most of the
attention that image tokens allocate to the text stream concentrates on these
template tokens rather than on the prompt tokens themselves. This pattern is not
merely an artifact of attention weights. Our causal interventions show that the
template tokens maintain object identity as \emph{implicit semantic registers}.
Consequently, semantic conditioning in deeper DiT layers relies primarily on
these emergent registers, rather than on image tokens repeatedly attending to
the original prompt tokens. Even more surprisingly, the template tokens do not
obtain their semantics by attending directly to the prompt. Instead, the first
block injects the prompt semantics into the image latents. In the second block,
the template tokens read this already-injected information back from the image
stream. Deeper image layers can then repeatedly attend to the template tokens
to acquire the semantics.
Text conditioning therefore follows an indirect circuit,
$\mathcal{S}\!\to\!\mathcal{I}\!\to\!\mathcal{R}$, rather than repeated direct
injection from $\mathcal{S}$ at every layer.

This circuit further reveals a division of labor across the heads and depths
inside the DiT. Across
heads, prompt-reading heads are largely causally inert, whereas register heads
preserve identity and a separate set of image-to-image heads performs visual
rendering. Across depth, early blocks commit the identity, middle blocks carry
it forward, and late blocks refine the result. These findings also
have a practical consequence. Because the prompt-reading heads are causally
inert, ranking heads by their image-to-semantic attention and pruning the
top-ranked ones yields a training-free, prompt-independent acceleration rule.
Pruning a quarter of all heads over the late denoising steps removes $20\%$ of
joint-attention FLOPs while GenEval~\citep{ghosh2023geneval} accuracy drops only $1.4$ points.

Overall, the main contributions of this paper are summarized as follows:
\begin{itemize}
\item We introduce a causal interpretability framework that traces how textual
information is injected, routed, stored, and transformed inside text-to-image
DiTs through token decomposition, span interventions, head transplantation, and
layer-wise masking.
\item We discover an indirect semantic-incorporation circuit: prompt semantics
enter the image stream in the first block and are then recovered and maintained
by nominally content-free chat-template tokens. These tokens act as robust
implicit semantic registers across models, timesteps, prompt complexity, and
language.
\item We uncover a division of labor across both heads and depth. Register heads
preserve object identity while image-to-image heads render visual structure.
Early layers commit identity, middle layers propagate it, and late layers refine
the output.
\item We translate this mechanistic account into a prompt-independent,
training-free pruning rule. Removing causally inert prompt-reading heads cuts
$20\%$ of joint-attention FLOPs while reducing GenEval accuracy by only $1.4$
points.
\end{itemize}

\section{Related Work}

This section reviews seminal studies on attention sinks and registers. An extended discussion is provided in Appendix~\ref{related}.

\paragraph{Attention Sinks in Generative Models.}
Sinks absorb disproportionate attention mass despite carrying little semantic content. First identified in autoregressive LLMs, where early tokens act as stable anchors whose cached states enable streaming inference~\citep{xiao2024efficient}, they are later tied to massive activations, depth-wise mixing control, and training dynamics and shown to emerge during pre-training rather than at initialization~\citep{sun2024massive,barbero2025llms,gu2025attention}. Similar content-agnostic concentration appears in multimodal models~\citep{kang2025see,yoo2026nature} and diffusion language models~\citep{rulli2025attention,dai2026revealing}. Such multimodal sink tokens may be removable or global-signal carriers depending on the model. Evidence in visual generation remains limited, and concurrent and independent work examines SD3/SDXL sinks~\citep{wu2026attention}. We instead study large-scale diffusion transformers with LLM-based text encoders and identify chat-template tokens as the dominant sinks that act as implicit semantic registers.

\paragraph{Registers as Attention and Computation Buffers.}
A complementary line of work controls where sinks reside via dedicated content-free slots. In vision transformers, register tokens absorb high-norm artifacts and internal computation, yielding cleaner feature or attention maps, with test-time variants that redirect them without retraining~\citep{darcet2024vision,jiang2026vision}, and in language models dedicated sink tokens likewise concentrate diffuse sink behavior and stabilize streaming inference~\citep{xiao2024efficient}, framing registers as explicit, steerable buffers. Whether this view transfers to diffusion transformers is only beginning to be studied. Added register tokens improve pixel-space DiTs as norm sinks and global-information carriers~\citep{starodubcev2026registers}, and some text-to-image DiTs such as FLUX-Schnell~\citep{flux2024} and PixArt-$\Sigma$~\citep{chen2024pixart,jamal2025diffusion} develop passive image-stream sink registers. We instead show that pre-existing chat-template tokens act both as attention registers and as implicit participants in semantic conditioning during generation, a role absent from the traditional register view.

\section{Preliminaries}
\label{sec:prelim}

\subsection{Text-to-Image Diffusion Transformers}
\label{sec:prelim-dit}

Text-to-image generators synthesize an image by iteratively denoising a latent under a flow-matching objective. Let $z \in \mathbb{R}^{n_{\mathrm{img}} \times d}$ be the sequence of $n_{\mathrm{img}}$ image latent tokens with timestep $t \in [0, 1]$, and let $c \in \mathbb{R}^{n_{\mathrm{cond}} \times d}$ denote the text conditioning. A transformer $f_\theta$ with $L$ blocks predicts the flow $f_\theta(z, t, c)$, which a sampler integrates from pure noise to a clean latent that is decoded to pixels by a VAE. We primarily study the publicly available Qwen-Image family~\citep{wu2025qwen}, a dual-stream multimodal diffusion transformer (MMDiT~\citep{esser2024scaling}) in which text and image tokens are processed by separate per-modality parameters within each block, yet interact through a single joint attention stream. Our analysis applies both to the multi-step base model and to its few-step distilled variant. Results on other models are reported in Appendix~\ref{extended}.

\subsection{Chat-Templated Text Conditioning}
\label{sec:prelim-cond}

In contrast to CLIP~\citep{radford2021learning}/T5-based~\citep{raffel2020exploring} models that include FLUX~\citep{flux2024}, SD3~\citep{esser2024scaling}, and SDXL~\citep{podell2024sdxl}, Qwen-Image obtains $c$ from a large vision--language model (Qwen2.5-VL~\citep{Qwen2.5-VL}) applied to a chat-formatted prompt. The user prompt $p$ is wrapped as
\[
c_{\mathrm{raw}}=[\,r_{\mathrm{prefix}};~p;~r_{\mathrm{suffix}}\,] \ ,
\]
where $r_{\mathrm{prefix}}$ contains the fixed system message and user-role markers, while $r_{\mathrm{suffix}}$ contains the closing user delimiter and assistant-role marker,
and $c$ is read from the encoder's last-layer hidden states after discarding the fixed template prefix. The complete template is given in Appendix~\ref{template}. The retained sequence $c=(c_1,\dots,c_{n_{\mathrm{cond}}})$ therefore splits into two disjoint, contiguous parts,
\[ c \;=\; \Big[\, \underbrace{c_1,\dots,c_{|\mathcal{S}|}}_{\text{semantic span }\mathcal{S}}\,,\;\; \underbrace{c_{|\mathcal{S}|+1},\dots,c_{n_{\mathrm{cond}}}}_{\text{template span }\mathcal{R}} \,\Big] \ , \]
where the semantic span $\mathcal{S}$ holds the prompt-content tokens and the template span $\mathcal{R}$ holds the trailing chat-template tokens that delimit the dialogue format and carry no prompt-specific content. These tokens arise solely from chat-template conditioning and have no counterpart in CLIP/T5 text encoders; they are the central object of our study.

\subsection{Joint Text--Image Attention}
\label{sec:prelim-attn}

\begin{wrapfigure}{r}{0.45\textwidth}
\vspace{-0.4em}
\centering
\makebox[\linewidth][c]{%
    \hspace*{-10pt}%
    \includegraphics[width=0.4\textwidth]{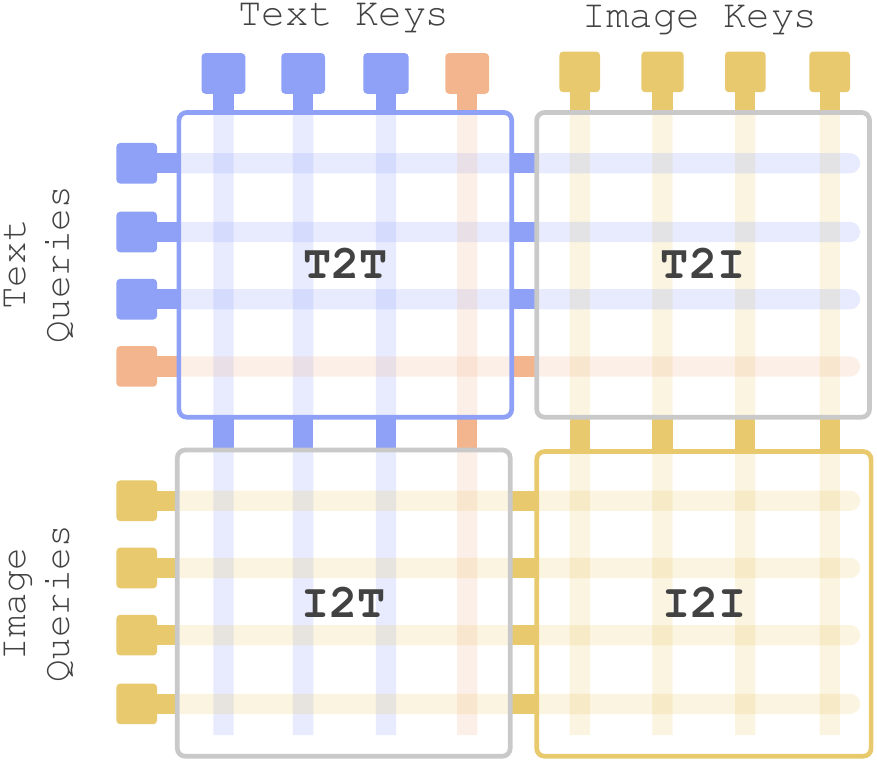}%
  }
\caption{Joint attention partition in MMDiT. Rows are queries and columns are keys, yielding T2T, T2I, I2T, and I2I regions over concatenated text and image tokens. We mainly focus on the I2T block to measure the attention mass received by text keys, especially the template tokens.}
\label{fig:fig_t}
\vspace{-2.5em}
\end{wrapfigure}

Within each MMDiT block, text and image tokens are processed by a single self-attention over the concatenation $[\,c\,;\,z\,]$, so the image stream attends to the conditioning tokens. As shown in Fig.~\ref{fig:fig_t}, partitioning the rows (queries) and columns (keys) of the joint attention matrix into their text and image segments yields four blocks: text queries attending to text and image keys (T2T and T2I) and image queries attending to text and image keys (I2T and I2I), the same decomposition used in recent analyses of MMDiT attention~\citep{shin2025exploring,helbling2025conceptattention,lv2025rethinking}. Among these, the I2T block is the cross-modal pathway through which the text conditioning $c$ acts on the image latents~\citep{shin2025exploring,lv2025rethinking}. We therefore focus our analysis on this block. To quantify how much attention each token---and in particular each template token in $\mathcal{R}$---receives, we work with the full (post-softmax) joint attention matrix at block $l$ and head $h$, $A^{(l,h)}\in\mathbb{R}^{(n_{\mathrm{cond}}+n_{\mathrm{img}})\times(n_{\mathrm{cond}}+n_{\mathrm{img}})}$, whose entry $A^{(l,h)}_{i,j}$ is the attention weight from query position $i$ to key position $j$ over the joint sequence $[\,c\,;\,z\,]$ (so each row sums to one). For a query group $\mathcal{Q}$ and a key span $\mathcal{K}$, we define the attention mass on $\mathcal{K}$ from $\mathcal{Q}$ as
\[
\begin{aligned}
m^{(l,h)}_{\mathcal{K}}(\mathcal{Q})
  &= \frac{1}{|\mathcal{Q}|}\sum_{i\in\mathcal{Q}}\sum_{j\in\mathcal{K}} A^{(l,h)}_{i,j} \ , \\
m^{(l,h)}_{j}(\mathcal{Q})
  &:= m^{(l,h)}_{\{j\}}(\mathcal{Q}) \ .
\end{aligned}
\]
with the subscript naming the key span that absorbs the mass and the argument the query group that emits it. We focus on the I2T block, i.e., image queries $\mathcal{Q}=\mathcal{I}$ (with $|\mathcal{I}|=n_{\mathrm{img}}$) attending to the conditioning keys $\mathcal{T}=\mathcal{S}\cup\mathcal{R}$, and abbreviate $m^{(l,h)}_{j}:=m^{(l,h)}_{j}(\mathcal{I})$ and $m^{(l,h)}_{\mathcal{K}}:=m^{(l,h)}_{\mathcal{K}}(\mathcal{I})$, so that $m^{(l,h)}_{\mathcal{R}}$ and $m^{(l,h)}_{\mathcal{S}}$ are the mass absorbed by the template and semantic spans, aggregated over heads, blocks, or timesteps where stated. Since $\mathcal{S}$ and $\mathcal{R}$ partition the text keys and the softmax is taken over the joint sequence, $m^{(l,h)}_{\mathcal{S}} + m^{(l,h)}_{\mathcal{R}}$ is the total mass of the I2T block, i.e.\ the fraction of I2T attention, and is in general below one, the remainder falling on the I2I block.

\begin{table*}[t]
\centering
\small
\setlength{\tabcolsep}{3.5pt}
\begin{tabular}{llcccccccc}
\toprule
Benchmarks & Models & $\overline{|\mathcal S|}$ &
$\overline{m}_{\mathcal R}$ & $\overline{m}_{\mathcal S}$ &
$\widetilde{m}_{\mathcal R}$ & $\widetilde{m}_{\mathcal S}$ &
$\frac{m_{\mathcal R}}{m_{\mathcal R}+m_{\mathcal S}}$ &
$\overline{m}_{\mathcal R}^{\mathrm{tok}} / \overline{m}_{\mathcal S}^{\mathrm{tok}}$ &
$\Pr[m_{\mathcal R} > m_{\mathcal S}]$ \\
\midrule
\multirow{2}{*}{GenEval} & Qwen-Image & \multirow{2}{*}{7.9} & 0.193 & 0.046 & 0.146 & 0.027 & 0.761 & 6.4 & 0.892 \\
    & Qwen-Image-2512 &  & 0.187 & 0.042 & 0.137 & 0.022 & 0.776 & 6.9 & 0.899 \\
\midrule
\multirow{2}{*}{DPG-Bench} & Qwen-Image & \multirow{2}{*}{82.1} & 0.076 & 0.169 & 0.046 & 0.140 & 0.292 & 7.2 & 0.123 \\
    & Qwen-Image-2512 &  & 0.081 & 0.172 & 0.047 & 0.143 & 0.291 & 7.5 & 0.122 \\
\midrule
\multirow{2}{*}{Qwen-Image-Bench} & Qwen-Image & \multirow{2}{*}{60.5} & 0.092 & 0.149 & 0.058 & 0.122 & 0.361 & 6.2 & 0.221 \\
    & Qwen-Image-2512 &  & 0.095 & 0.149 & 0.059 & 0.122 & 0.360 & 6.5 & 0.223 \\
\bottomrule
\end{tabular}
\caption{
Attention statistics across prompt benchmarks, in which $\overline{|\mathcal S|}$ is the mean number of semantic tokens per prompt, $m_{\mathcal R}$ denotes image-to-text attention mass assigned to trailing template tokens, $m_{\mathcal S}$ denotes mass assigned to semantic prompt tokens, $\overline{m}_{\mathcal R}^{\mathrm{tok}} / \overline{m}_{\mathcal S}^{\mathrm{tok}}$ denotes the average mass per template token divided by the average mass per semantic token, and $\Pr[m_{\mathcal R} > m_{\mathcal S}]$ is the fraction of step--layer--head sites where template mass exceeds semantic mass. Bars denote means and tildes denote medians, averaged over prompt-level statistics.
}
\label{tab:benchmark_attention}
\end{table*}

\section{Text Template Tokens as Implicit Semantic Registers}

In this section, we introduce our causal interpretability framework and apply
it to show that the chat-template tokens act as the model's implicit semantic
registers. Combining token-level attention decomposition with span-level
conditioning interventions, cross-trajectory head transplantation, and
layer-wise causal masking, it traces where conditioning information is read,
routed, and stored during denoising. Building on this analysis, we further
characterize distinct generative mechanisms across heads and depths and derive
a training-free head-pruning rule. The following subsections develop
these findings in turn.

More analyses, generalization to other models, and additional qualitative
examples are provided in Appendix~\ref{extended}.

\subsection{Template Tokens Are Dominant Attention Sinks}
\label{sec:sinks}

We begin with a purely descriptive question: across the denoising trajectory, where does the image stream actually direct its attention over the conditioning tokens? Using the I2T mass defined in Preliminaries, we aggregate $m^{(l,h)}_{\mathcal{K}}$ over the $L$ blocks, $H$ heads, and $T$ sampled timesteps, writing $\bar m_{\mathcal{K}} = \tfrac{1}{LHT}\sum_{l,h,t} m^{(l,h)}_{\mathcal{K}}(t)$ for the mean mass absorbed by a key span $\mathcal{K}$.

\begin{figure*}[t]
\centering
\includegraphics[width=0.95\textwidth]{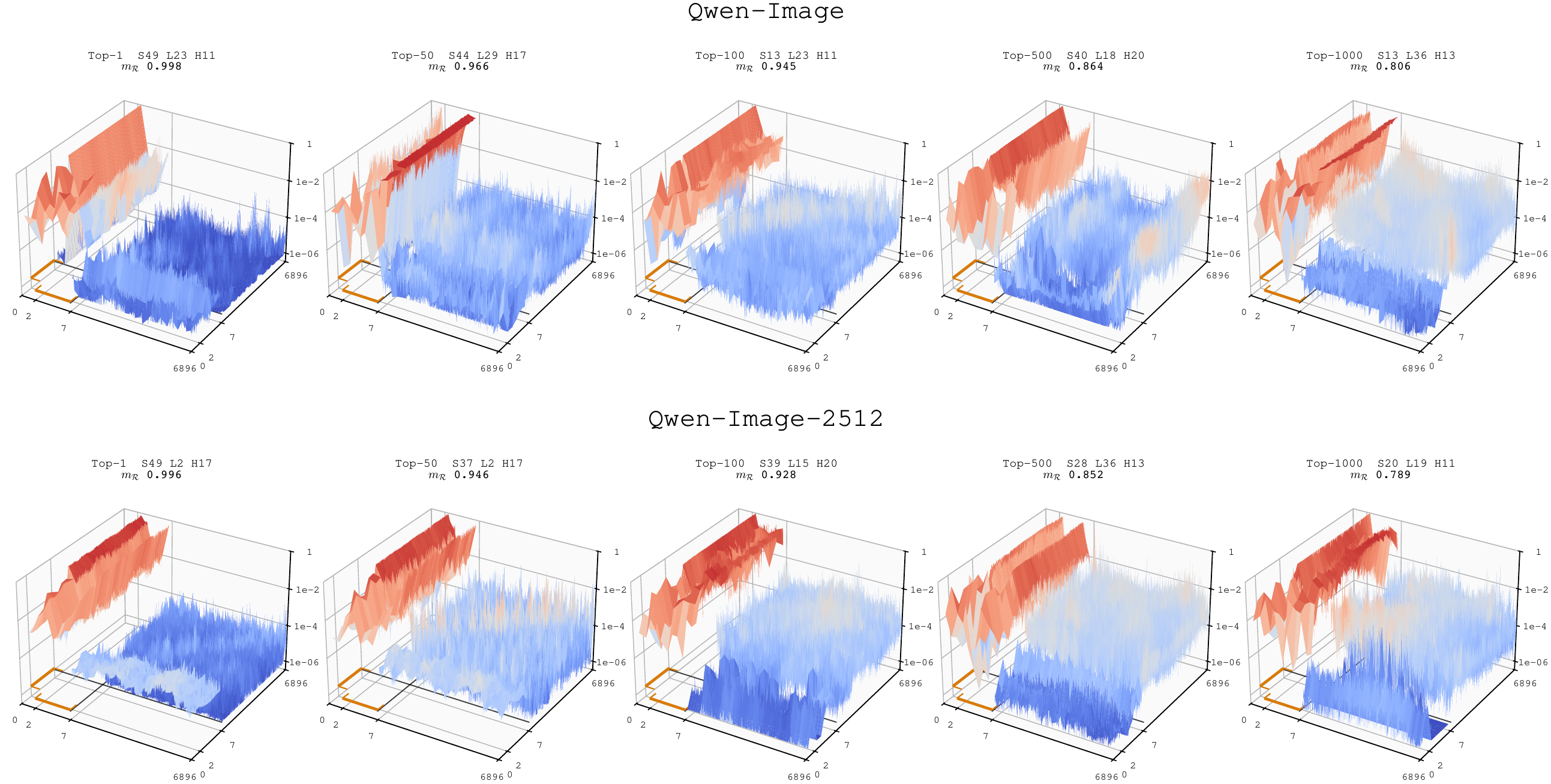}
\caption{
3D surfaces of full query--key attention maps for the prompt ``An apple'' on Qwen-Image and Qwen-Image-2512.
Columns show Top-$n$ heads ranked by $m_{\mathcal R}$, the image-to-template-token attention mass. Attention concentrates as a sharp ridge over the template span $\mathcal R$ while attention among image tokens stays low.
The text and image token ranges are displayed with a nonuniform axis scale for readability, and orange markers denote the template token span $\mathcal R$.
}
\label{fig:head3d}
\end{figure*}

\paragraph{A minimal example.}
We start from the simplest possible prompt, ``An apple'', whose retained
conditioning sequence consists of only $|\mathcal{S}|{=}2$ content tokens
followed by the $|\mathcal{R}|{=}5$ template tokens. On both Qwen-Image and
Qwen-Image-2512, image queries place an order of
magnitude more attention on the content-free template span than on the prompt
content. Fig.~\ref{fig:head3d} visualizes this for the top sink heads (those
with the largest $m_{\mathcal{R}}$). The attention surface forms a sharp ridge
over the template span $\mathcal{R}$ while attention among image tokens stays
low, with the top head reaching $m_{\mathcal{R}}=0.998$.
On Qwen-Image,
$\bar m_{\mathcal{R}} = 0.23$ versus $\bar m_{\mathcal{S}} = 0.020$---an
$11\times$ span-level gap, or $4.6\times$ per
token---and the template span outweighs the semantic span at $98\%$ of all
$(t,l,h)$ sites. The mass collapses onto a few delimiter
tokens: \texttt{<|im\_end|>} alone absorbs $13\%$ of all
image-query attention at the highest noise level---more than $6\times$ the
content noun ``\,apple''---and is the dominant sink at every sampled timestep.
Qwen-Image-2512 behaves nearly identically.
The template span thus behaves as a classical attention
sink~\citep{xiao2024efficient}: a few positions absorbing a disproportionate
share of attention while carrying no prompt-specific content.

\paragraph{At scale across prompts.}
To show this is not an artifact of a single toy prompt, we measure the same
quantities at scale over three benchmarks spanning two axes along which the sink
could plausibly weaken: prompt complexity and language. Along the first axis the
prompts grow in length and compositional density, from the minimal ``An apple''
($|\mathcal S|{=}2$), to the moderately complex, object-focused prompts of
GenEval~\citep{ghosh2023geneval} ($553$ prompts, mean $7.9$ content tokens), to
the highly complex, long and dense multi-object prompts of
DPG-Bench~\citep{hu2024ella} ($1{,}065$ prompts, mean $82.1$ content tokens).
The second axis is covered by Qwen-Image-Bench~\citep{li2026qwen} ($1{,}000$
Chinese prompts, mean $60.5$ content tokens). We evaluate each on both checkpoints
and report Qwen-Image numbers in the text, with Qwen-Image-2512 closely matching
throughout (Table~\ref{tab:benchmark_attention}). On GenEval the picture is unchanged.
The five template tokens absorb
$\bar m_{\mathcal{R}} = 0.19$ of all image-query attention versus
$\bar m_{\mathcal{S}} = 0.046$ for the content tokens, a per-token
ratio of $6.4\times$. Equivalently, $76\%$ of all
I2T attention lands on the template span. The dominance holds
site by site: the template span outweighs the entire semantic span at
$89\%$ of $(t,l,h)$ sites, and a template token is the single
most-attended text token at $91\%$ of sites, with \texttt{<|im\_end|>}
alone the top sink at $54\%$. The smaller span-level ratio ($\sim\!4\times$
versus $11\times$ for ``An apple'') merely reflects longer prompts spreading
the semantic mass, while the per-token gap, which controls for length, is stable.
Pushing further along the complexity axis, the $10\times$ longer prompts of
DPG-Bench invert the span-level mass as expected ($\Pr$ drops to $12\%$), yet
the per-token ratio rises to $7.2\times$, so the sink strengthens rather than
weakens with prompt length. The Chinese prompts of Qwen-Image-Bench behave the
same ($\Pr=22\%$, per-token ratio $6.2\times$), with \texttt{<|im\_end|>} still
the single largest sink. The template sink is thus robust to both prompt
complexity and language.

\subsection{Template Tokens Carry Little Semantics}

\begin{figure*}[t]
\centering
\includegraphics[width=0.95\textwidth]{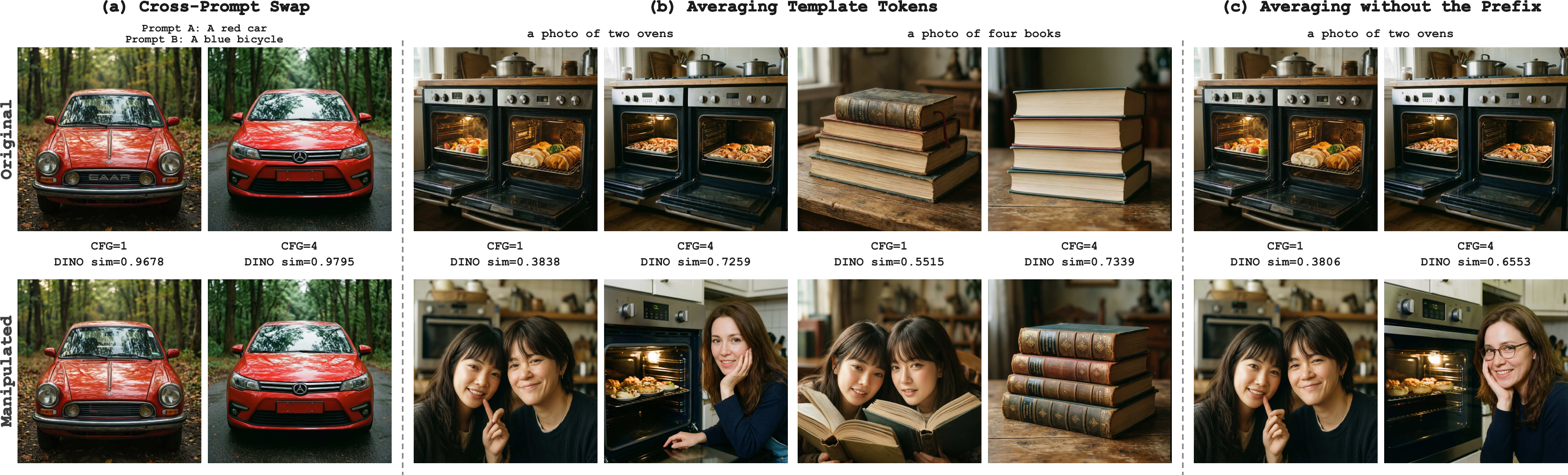}
\caption{Template tokens carry semantics, but only weakly. (a) Cross-prompt swap substitutes another prompt's template tokens, $[\mathcal{S}^{A};\mathcal{R}^{B}]$, with little effect on the object. (b) Average on template tokens replaces $\mathcal{R}$ with a single prompt-agnostic average $\bar{\mathcal{R}}$, $[\mathcal{S};\bar{\mathcal{R}}]$, which can occasionally cause large semantic changes. Increasing the guidance scale (CFG) largely resolves these deviations. (c) Recomputing the average $\bar{\mathcal{R}}$ from the same prompts with the ``a photo of'' prefix stripped leaves the result essentially unchanged from (b).}
\label{fig:swapmean}
\end{figure*}

Since the trailing template tokens $\mathcal{R}$ attend to the preceding prompt
tokens $\mathcal{S}$ in the encoder, do they carry the prompt's semantics? We
probe this with a cross-prompt swap. For a pair of prompts $A,B$ we build, at
fixed seed and sampler, a template-token swap $[\,\mathcal{S}^{A};\mathcal{R}^{B}\,]$
that keeps $A$'s content tokens but substitutes $B$'s template tokens. If template
tokens carried semantics, the template-token swap would drift toward $B$.
Fig.~\hyperref[fig:swapmean]{\ref*{fig:swapmean}a} shows it does not: the template-token swap
preserves $A$'s semantics, with only minor changes to prompt-irrelevant
details, as confirmed by the DINOv3 \citep{simeoni2025dinov3} \texttt{[CLS]} cosine
similarity $\mathrm{sim}(\cdot,\cdot)$ between the template-token swap and $A$.

To test whether $\mathcal{R}$ holds any prompt-specific content at all, we
average $\mathcal{R}$ over $100$ GenEval prompts into a single prompt-agnostic
template $\bar{\mathcal{R}}$, then substitute it for the template span of each
of $40$ disjoint held-out prompts, forming $[\,\mathcal{S};\bar{\mathcal{R}}\,]$. Although this changes the template tokens'
representation by a relative $L_2$ of $0.36$ and perturbs the model's first
denoising-step prediction by $1.8\%$, the
object is essentially unchanged. On Qwen-Image, the median own-vs-average image
similarity is $0.99$ and the mean is $0.93$, so a single content-free template shared
across all prompts suffices to render any object. The exceptions are instructive: among the
evaluation prompts below $\mathrm{sim}{=}0.90$, most keep the object, count, and
attributes, differing only in prompt-irrelevant content, while a few alter the object, count, or attributes themselves.
As shown in Fig.~\hyperref[fig:swapmean]{\ref*{fig:swapmean}b}, many of these
deviations are drastic. The shared average template can replace the object
entirely, change its count, or drop key attributes. With classifier-free guidance~\citep{ho2022classifier}, however, these deviations are largely
suppressed and the intended object is recovered. Together, these experiments show that template tokens carry semantics, but only weakly.

A natural question is why the genuine failures collapse to a generic human
portrait rather than to noise or to some other object. Generating from an empty prompt
reproduces the same outcome, so the portrait is the model's unconditional
default, presumably reflecting the predominance of people in photo--caption
training data. Based on this observation, we recompute the averaged template $\bar{\mathcal{R}}$
from the same prompts with the ``a photo of'' prefix stripped and regenerate.
The output is unchanged (Fig.~\hyperref[fig:swapmean]{\ref*{fig:swapmean}c}).
This shows that the semantic content carried by the template tokens is almost
negligible. This collapse is therefore better understood as a thresholding effect: when the semantics
are too weak to assert the target at low guidance, generation relaxes to this
default.

\subsection{Semantics Reside in Template Registers Implicitly}
\label{sec:register}

Attention sinks are usually taken to be content-free, so one would expect the generated
object to be determined by the semantic span $\mathcal{S}$ rather than by $\mathcal{R}$.
Indeed, in the previous section we found that the VLM-encoded template-token latents
carry little semantic information. However, we find that inside the DiT it is in fact the
template tokens $\mathcal{R}$ that act as the semantic registers. We also trace the
process by which semantics enter these tokens, and find that the flow is implicit:
semantics pass first from the semantic tokens $\mathcal{S}$ into the image tokens, and only
then from the image tokens into the template tokens $\mathcal{R}$.
Throughout, we use the public $2$-step distilled LoRA checkpoint from
Wuli-art~\citep{li2026turbolora} on Qwen-Image-2512, but run only a single denoising step,
so that the timestep is fixed and its effect can be ignored, keeping the experimental
design and analysis simple.

\begin{figure*}[t]
\centering
\includegraphics[width=0.92\textwidth]{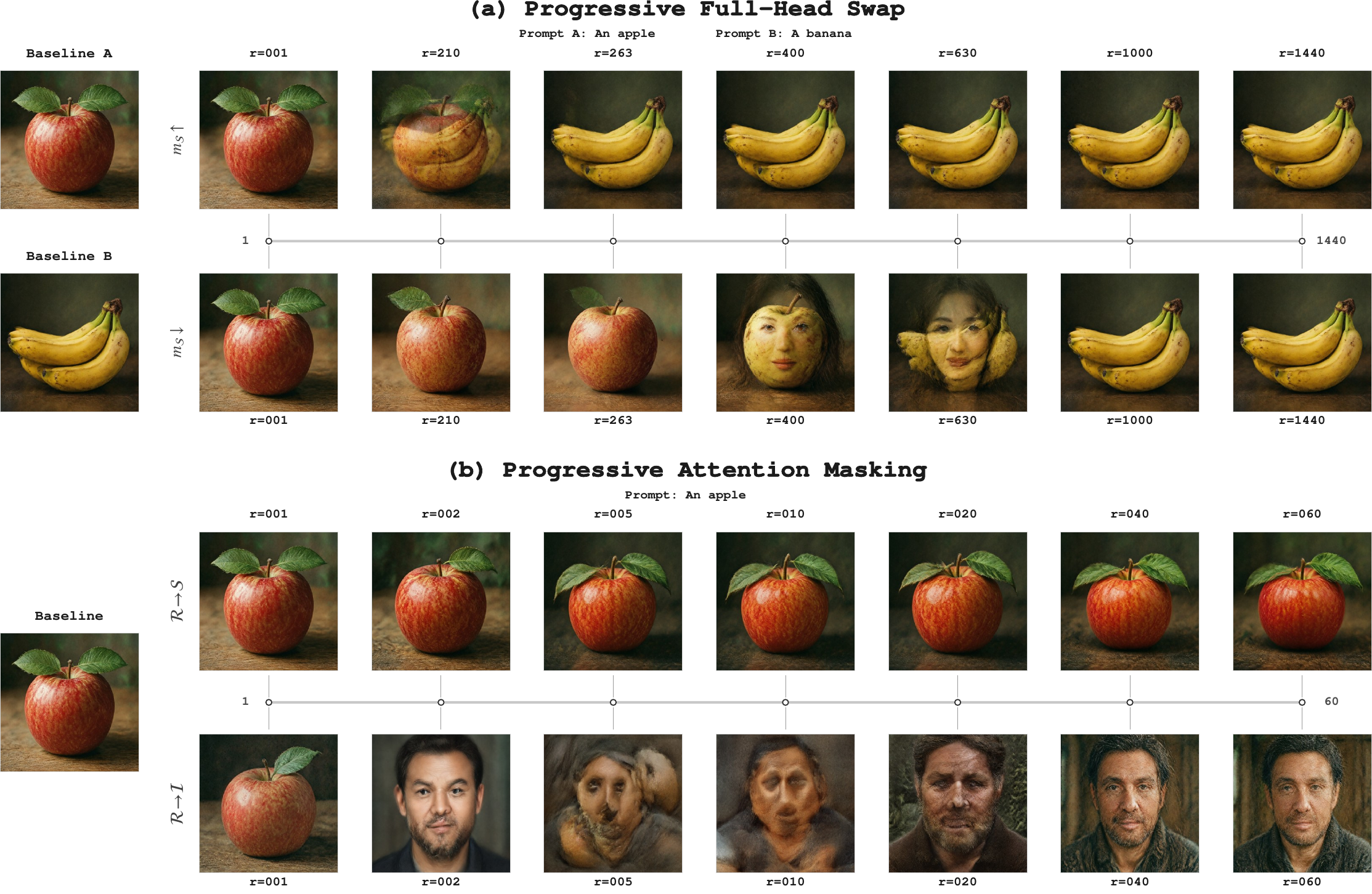}
\caption{Template tokens act as the implicit semantic registers in the denoising process. (a) Progressive head swap, ordering heads by their semantic-span attention $m_\mathcal{S}$. Swapping high-to-low ($m_\mathcal{S}{\downarrow}$, bottom) barely transfers the identity and even collapses to the unconditional portrait; the reverse order ($m_\mathcal{S}{\uparrow}$, top) flips apple to banana after only ${\sim}18\%$ of the heads. (b) Progressive attention masking over layers. Blocking $\mathcal{R}{\to}\mathcal{S}$ attention leaves the apple largely intact, while blocking $\mathcal{R}{\to}\mathcal{I}$ attention quickly erases the object identity.}
\label{fig:semantic}
\end{figure*}

\paragraph{Template tokens are semantic registers.}
We probe the computation with a cross-trajectory transplant. We run two
denoising trajectories, one for $A{=}$``An apple'' and one for $B{=}$``A banana'', and
progressively copy per-head attention projections $(q,k,v)$ from the $B$ run into the $A$
run. At round $n$ we replace, in $A$, the $(q,k,v)$ of $n$ of the $L{\times}H{=}1{,}440$
attention heads with their $B$ counterparts.
Under the conventional ``semantic-head'' view, the object should live in the heads that
most read the semantic tokens, so transplanting those first ought to flip the identity
soonest. Fig.~\hyperref[fig:semantic]{\ref*{fig:semantic}a} tells the opposite story: how much a head reads $\mathcal{S}$ is a poor indicator of its causal role. Transplanting the top semantic readers first fails to transfer the target identity and instead pulls generation toward the unconditional default. The identity turns over only once the heads that barely attend to $\mathcal{S}$ are swapped in. In this reverse order, just ${\sim}18\%$ of the $1{,}440$ heads suffice to flip the apple into a banana. Reading the prompt and carrying the object are thus decoupled, our first indication that the semantics reside in the template registers $\mathcal{R}$ rather than in the span that is read.

\paragraph{Semantics enter the registers implicitly.}
We next ask how the registers acquire their object content. In
the joint attention we mask, at the register queries $\mathcal{R}$, their attention to a
single key span---either the semantic tokens $\mathcal{S}$ or the image latents---before the
softmax, applied cumulatively over the first $k$ blocks. The two spans
behave oppositely (Fig.~\hyperref[fig:semantic]{\ref*{fig:semantic}b}). Cutting $\mathcal{R}$'s attention to $\mathcal{S}$ leaves the object
essentially unchanged. The
register does not obtain its content by reading the prompt tokens directly. Cutting
$\mathcal{R}$'s attention to the image latents instead collapses the object within the first two blocks. The register therefore
assembles its content by attending to the image stream, and does so early in the
network---not from the prompt span, and not from its own near-content-free embedding.
We further note that masking the first block alone has little effect. Since the image latents there are still pure Gaussian noise and carry no object content, this indicates that the prompt-to-image injection of semantics ($\mathcal{S}{\to}\mathcal{I}$) completes within the first block.
And the image-to-template injection ($\mathcal{I}{\to}\mathcal{R}$) completes in the second block.

\subsection{Distinct Mechanisms at Different Heads and Different Depths}
\label{sec:depth}

In visual generation DiTs, what role the different heads and the layers at
different depths play during generation has long remained unclear. Building on our
head-wise analysis, we examine both and uncover a division of labor.

\paragraph{Different heads.}
The heads themselves fall into two functionally distinct groups. The first are the
semantic heads identified above, the register heads that encode which object to
generate and hold that identity across the denoising trajectory.
The second are rendering heads, whose
attention concentrates almost entirely in the I2I block.
They hold no object identity but instead knit the image latents into a spatially
coherent picture. Ablating even a small fraction of them collapses the entire
output into a mosaic of incoherent patches (Fig.~\ref{fig:render}).
Note that the division is not absolute---a few heads blend these roles to some
degree---but it remains clear overall.

\paragraph{Different depths.}
To examine how identity processing changes with depth, we localize the ${\sim}270$ heads that most
drive the identity transfer (top of the $m_\mathcal{S}{\uparrow}$ order) across the
$60$ transformer blocks (Fig.~\ref{fig:layer}). Their distribution is
bimodal---$89$ heads in the early blocks (L1--10), $85$ in the late blocks
(L51--60), and only $96$ thinly spread across the forty middle blocks
(L11--50)---and restricting the swap to each group exposes a matching division of
labor. The early heads alone flip the object cleanly in both directions, so
identity is committed early, consistent with the $\mathcal{S}{\to}\mathcal{I}$
injection completing in the first block. The middle heads alone leave it almost
unchanged, carrying the identity forward rather than rewriting it. And the late
heads alone flip it only with residual distortion, refining rather than deciding
it. This three-stage organization---early commit, middle carry, late
refine---mirrors that reported in LLMs~\citep{queipo2025attention,lad2026remarkable}.

\subsection{Semantic Registers Guide Training-Free Head Pruning}

Our analysis translates directly into a training-free acceleration rule. Because
heads with high image-to-semantic attention $m_\mathcal{S}$ are causally inert,
with transplanting them first failing to transfer object identity, we rank the
$L{\times}H{=}1{,}440$ heads by descending $\bar m_{\mathcal S}$ and, for a budget
$K$, prune the computations of the top-$K$ heads. Pruning a head removes its full
per-head cost, namely the $q,k,v$ and output projections together with the
attention. In practice we prune only over the last $80\%$ of the denoising steps,
since the early steps are more sensitive and commit the object's identity. Every
component of the joint-attention block scales linearly with the number of heads,
so pruning $K$ of the $1{,}440$ heads over the late fraction $1{-}f$ of the
schedule removes a fraction $(1{-}f)\,K/1440$ of its FLOPs, with $f{=}0.2$. The rule is prompt-independent and requires no per-prompt profiling or runtime head selection.

\begin{wraptable}{r}{0.5\textwidth}
\vspace{-1.0em}
\centering
\small
\setlength{\tabcolsep}{4pt}
\begin{tabular}{lcccc}
\toprule
$K$ & GenEval & LPIPS & HPSv3 & FLOP $\downarrow$ \\
\midrule
$0$   & 76.1 & --   & 9.56 & $0\%$ \\
$216$ & 75.8 & 0.21 & 9.47 & $12.0\%$ \\
$288$ & 75.5 & 0.29 & 9.29 & $16.0\%$ \\
$360$ & 74.7 & 0.39 & 8.76 & $20.0\%$ \\
$400$ & 74.0 & 0.40 & 8.44 & $22.2\%$ \\
\bottomrule
\end{tabular}
\caption{
Training-free late-step head pruning on Qwen-Image-2512, evaluated on GenEval.
$K$ is the number of silenced heads out of $1{,}440$, while $K{=}0$ is
the unpruned baseline.
}
\label{tab:accel}
\vspace{-1.4em}
\end{wraptable}

Table~\ref{tab:accel} reports GenEval accuracy on
Qwen-Image-2512 across silencing budgets, alongside LPIPS~\citep{zhang2018unreasonable} and the
HPSv3~\citep{ma2025hpsv3} preference score. Object correctness is well preserved. Pruning
a quarter of all heads ($K{=}360$) removes $20\%$ of the joint-attention
FLOPs while GenEval accuracy drops only $1.4$ points, from $76.1$ to $74.7$.
Perceptual quality declines more quickly, so a lighter $K{=}216$ gives
the best trade-off. What matters is which heads, not how many: at $K{=}288$,
GenEval falls from $75.5$ under our $\bar m_{\mathcal S}$ ranking to $69.6$ under
the template-sink ranking $\bar m_{\mathcal R}$, whose register heads are more
load-bearing, and to $51.3$ under random selection (HPSv3 $9.29$, $8.03$, $4.86$).

\section{Conclusions}

Despite rapid progress in diffusion transformers, how they actually read,
route, and transform information during generation remains poorly understood.
Our findings revise the conventional picture of text conditioning in diffusion
transformers. Prompt semantics do not simply remain attached to the tokens that
encode them, hinting that the model's internal representation is more complex than our current understanding.
Our proposed interpretability framework provides a systematic way to trace the causal flow of information through the model.
We believe that a deeper understanding of these internal mechanisms can inform the design of more capable and efficient generative models.

\bibliographystyle{plainnat}
\bibliography{references}

\newpage
\beginappendix

\begin{figure}[H]
\centering
\includegraphics[width=\textwidth]{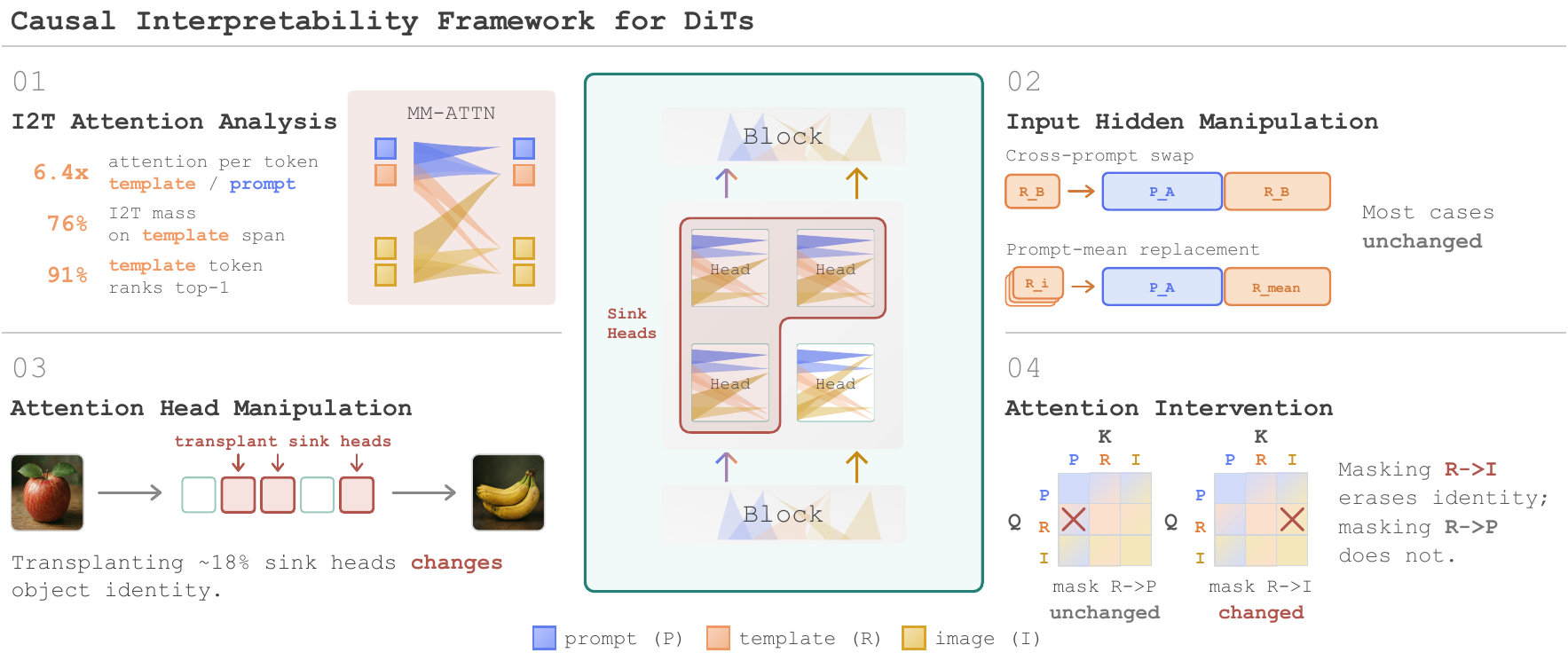}
\caption{Overview of our causal interpretability framework.}
\label{fig:framework}
\end{figure}

\section{Limitations and Future Work}

Our study is primarily analytical. The training-free head-pruning rule is only a
naive first-cut use of our analysis, and more sophisticated designs---for
instance, sink-aware sparse attention that keeps the register positions as keys
while pruning the rest---could yield substantially larger gains. Equally open is why these content-free template tokens come to
serve as registers in the first place, which we leave to future investigation.

\section{Qwen-Image Chat Templates and Retained Spans}
\label{template}

\begin{figure}[H]
\centering
\includegraphics[width=\textwidth]{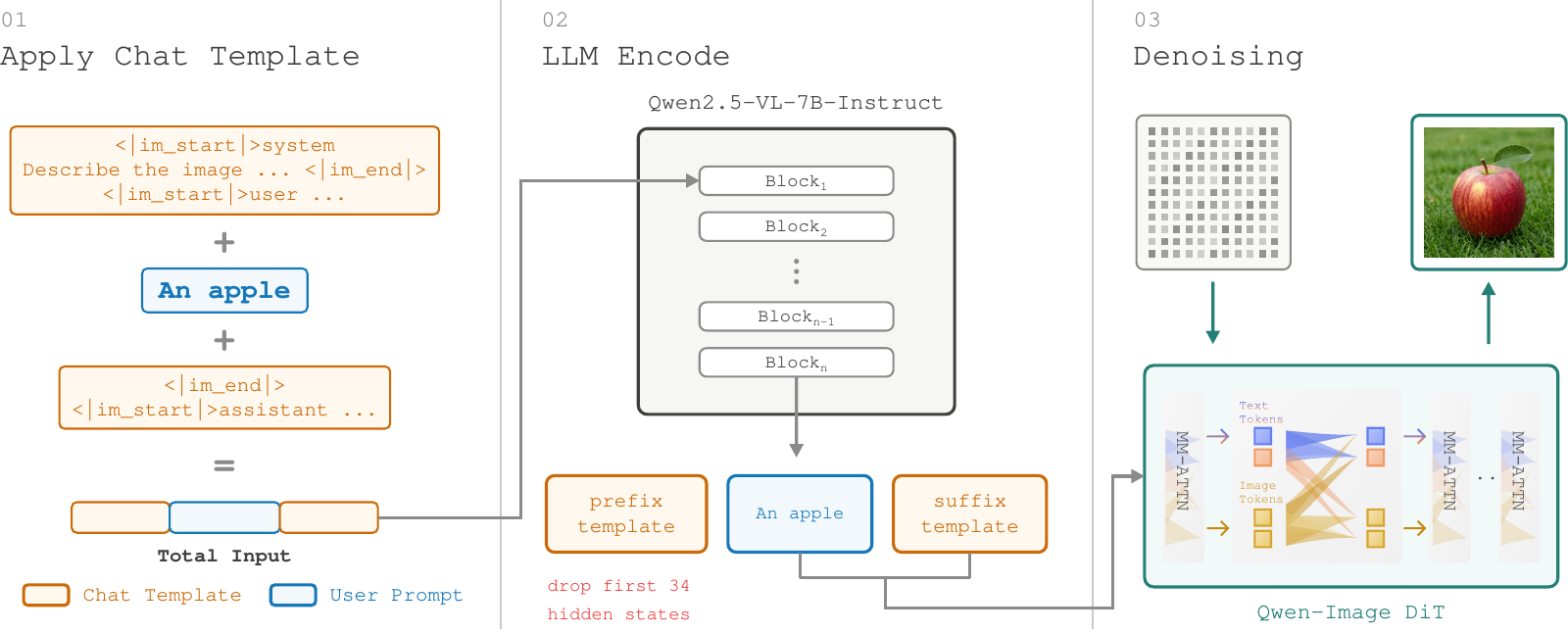}
\caption{Chat-templated text conditioning in Qwen-Image.}
\label{fig:qwenimage}
\end{figure}

We consider three \texttt{Qwen-Image} family checkpoints: the original \texttt{Qwen-Image}, the updated
\texttt{Qwen-Image-2512} text-to-image model, and the instruction-based editing model
\texttt{Qwen-Image-Edit-2511}. The first two use the same text-to-image chat template.
Fig.~\ref{fig:qwenimage} gives an overview of this chat-templated conditioning pipeline.
Line numbers in the bubbles mark literal newline positions.

\noindent\textit{Qwen-Image and Qwen-Image-2512.}\par
\begin{templatebubble}
<|im_start|>system
Describe the image by detailing the color, shape, size, texture, quantity, text, spatial relationships of the objects and background:<|im_end|>
<|im_start|>user
{prompt}<|im_end|>
<|im_start|>assistant

\end{templatebubble}

After preprocessing before entering the DiT.
\begin{templatebubble}
{prompt}<|im_end|>
<|im_start|>assistant

\end{templatebubble}

\noindent\textit{Qwen-Image-Edit-2511.}
This editing checkpoint uses a different template. The system message asks the encoder to describe
the input image and apply the user's edit instruction, while the user turn contains visual placeholder
tokens before the text prompt. For multiple input images, the pipeline prepends one visual placeholder
per image by looping over the image list and formatting it as \texttt{Picture i:}. In the bubble below,
the ellipsis abbreviates repeated \texttt{Picture i:} blocks and is not a literal template token.

\begin{templatebubble}
<|im_start|>system
Describe the key features of the input image (color, shape, size, texture, objects, background), then explain how the user's text instruction should alter or modify the image. Generate a new image that meets the user's requirements while maintaining consistency with the original input where appropriate.<|im_end|>
<|im_start|>user
Picture 1: <|vision_start|><|image_pad|><|vision_end|>Picture 2: <|vision_start|><|image_pad|><|vision_end|>...Picture N: <|vision_start|><|image_pad|><|vision_end|>{prompt}<|im_end|>
<|im_start|>assistant

\end{templatebubble}

After preprocessing before entering the DiT.
\begin{templatebubble}
Picture 1: <|vision_start|><|image_pad|><|vision_end|>Picture 2: <|vision_start|><|image_pad|><|vision_end|>...Picture N: <|vision_start|><|image_pad|><|vision_end|>{prompt}<|im_end|>
<|im_start|>assistant

\end{templatebubble}

\section{More Discussion on Related Work}
\label{related}

\subsection{Evolution of Diffusion Models}

Diffusion generative modeling evolved from denoising and score-based formulations that learn to reverse a gradual noising process into practical, high-fidelity image generators. DDPMs and score-based SDEs established the core denoising/score view, while DDIM sampling, learned variances, and guidance improved sampling efficiency and controllability~\citep{ho2020denoising,song2020score,song2020denoising,nichol2021improved,dhariwal2021diffusion,ho2022classifier}. A second shift moved generation into compressed latent spaces and attached text conditioning through pretrained encoders and cross-attention, enabling scalable text-to-image systems~\citep{rombach2022high,podell2024sdxl}. Recent architectures increasingly replace U-Net denoisers with transformer backbones, spanning ViT-style tokenization, scalable DiT blocks, masked-token training, weak-to-strong text-to-image scaling, and broader DiT architectural refinements~\citep{bao2023all,peebles2023scalable,gao2023mdtv2,chen2024pixart,xu2026rethinking}. This DiT line is often paired with flow-matching or rectified-flow objectives and has expanded to multimodal joint-attention streams~\citep{lipman2022flow,esser2024scaling,wu2025qwen}. This transition is important for attention-sink analysis because conditioning tokens are no longer only side information injected through cross-attention. In joint-attention DiTs, prompt tokens, structural template tokens, and image latents compete within the same attention computation.

\subsection{Attention Sinks in Sparse Attention}

Beyond mechanistic analyses, attention sinks have also motivated sink-aware sparse attention. In long-context LLMs, sink tokens and related high-mass positions are often preserved while low-mass tokens are evicted, yielding stable streaming or compressed KV caches~\citep{xiao2024efficient,zhang2023h2o,liu2023scissorhands,li2024snapkv}. Recent sparse-attention systems extend this idea from token retention to head specialization: local or streaming heads rely on recent tokens and attention sinks, while retrieval heads keep access to long-range context~\citep{xiao2025duoattention,tang2025razorattention,zhou2026full}. Complementary dynamic schemes instead choose sparse patterns or query-relevant token/page subsets through context-aware budgets or lightweight token indexers~\citep{jiang2024minference,lai2025flexprefill,tang2024quest,wang2026fasa}. This line of work treats sinks not merely as artifacts to be explained, but as useful attention anchors that can be preserved, isolated, or combined with retrieval to reduce inference cost.

\section{Experimental Setup}
\label{sec:setup}

Unless otherwise specified, all experiments use Qwen-Image-2512 as the
default model with a fixed random seed of $42$, and are implemented with
PyTorch 2.12.1 on top of the DiffSynth-Studio~\citep{diffsynthstudio} (v1.1.7)
framework. All runs are carried out on NVIDIA H20 GPUs with $1{,}024$GB of GPU
memory and a 192-core CPU host running CentOS 7.

\section{Extended Experiments}
\label{extended}

In this section, we provide additional results for the experiments presented in
the main text, along with further experiments not included there.

\subsection{Extending the Analysis to FLUX.2}

FLUX.2~\citep{flux2bfl} is a 32B rectified-flow DiT whose text conditioning is
produced by a Mistral-3 (24B) vision--language model. Its prompt is wrapped in a Mistral chat template with
\texttt{system} and \texttt{user} roles (with no assistant turn) and encoded by
the VLM. The complete template is given below. Unlike Qwen-Image, no fixed
prefix is discarded, so the entire serialized sequence is retained as the
conditioning $c$.

\begin{templatebubble}
<s>[SYSTEM_PROMPT]You are an AI that reasons about image descriptions. You give structured responses focusing on object relationships, object
attribution and actions without speculation.[/SYSTEM_PROMPT][INST]{prompt}[/INST]
\end{templatebubble}

The I2T sink of Sec.~\ref{sec:sinks} carries over as well. Aggregating
the I2T mass over all $553$ GenEval prompts, the $33$
retained template tokens of $\mathcal{R}$ absorb $\bar m_{\mathcal R}=0.22$ of all
image-query attention against $\bar m_{\mathcal S}=0.016$ for the mean $8.1$
content tokens, so $92\%$ of all I2T attention lands on the
template span, at a per-token ratio of $3.3\times$. The dominance is near-total
site by site, with the template span outweighing the entire semantic span at
$99.8\%$ of $(t,l,h)$ sites. The one difference from Qwen-Image is where the mass sits
within $\mathcal{R}$. Rather than collapsing onto a single delimiter, it spreads
across the system block, with tokens such as ``\,that'', \texttt{[/SYSTEM\_PROMPT]},
and \texttt{<s>} as the leading sinks and ``\,that'' alone the most-attended token
at $33\%$ of sites. Still, every dominant sink is a template token and no content token
appears among them, so the chat-template sink is intrinsic to FLUX.2 despite its
different encoder, template, and architecture.

\begin{figure}[!htbp]
\centering
\includegraphics[width=\textwidth]{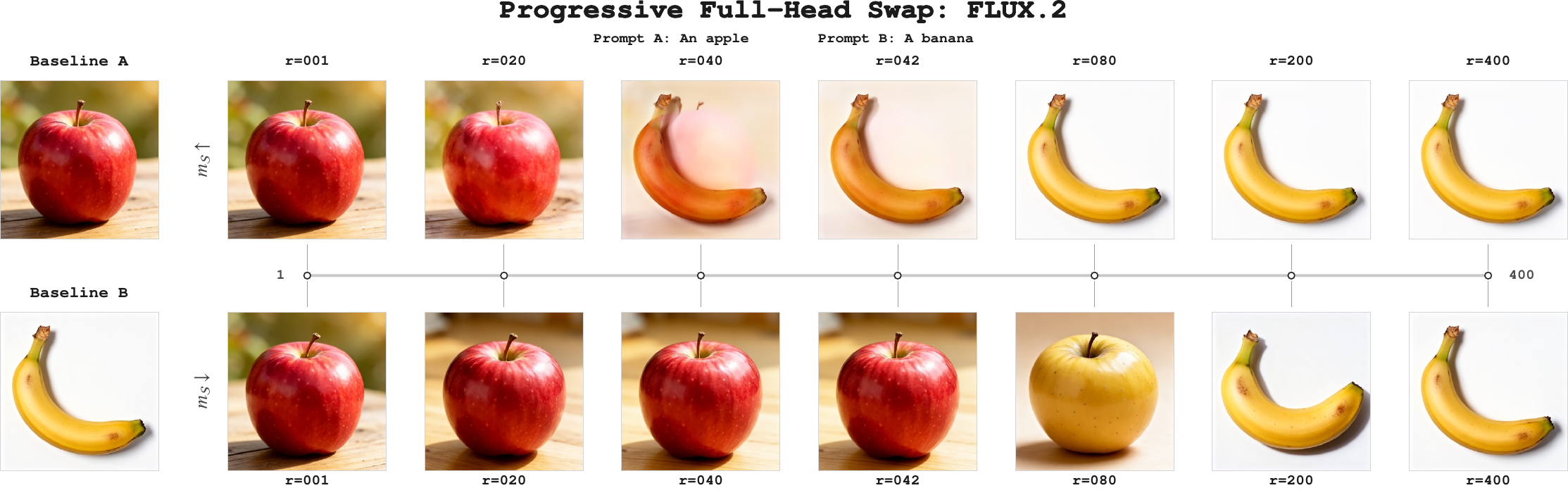}
\vspace{-0.8em}
\caption{Results for the progressive head-swap experiment on FLUX.2.}
\label{fig:flux2}
\end{figure}

On FLUX.2, the progressive head-swap experiment likewise supports our conclusion
(Fig.~\ref{fig:flux2}).

\subsection{Extending the Analysis to Krea-2-Turbo}

Krea-2-Turbo~\citep{krea2} is a 12B single-stream flow-matching DiT whose text
conditioning is produced by a Qwen3-VL~\citep{Qwen3-VL} (4B) vision--language model, a design close
to Qwen-Image. The template is identical to that of Qwen-Image
(Appendix~\ref{template}).

The template sink recurs here. Aggregating the I2T mass over blocks, heads, and
the model's $8$ sampling steps and averaging over the $553$ GenEval
prompts (mean $|\mathcal S|{=}7.9$), the template span absorbs
$\bar m_{\mathcal R}=0.11$ of the image-query attention versus
$\bar m_{\mathcal S}=0.053$ for the prompt content. Equivalently, $67\%$ of all I2T attention lands on
the template span, a per-token ratio of $3.1\times$. Site by site, the template
span outweighs the entire semantic span at $77\%$ of
$(t,l,h)$ sites, and a template token is the single most-attended text token at
$83\%$ of them. As on Qwen-Image the mass collapses onto one delimiter, though here
it is the trailing newline after \texttt{assistant} (dominant at $59\%$ of sites)
rather than \texttt{<|im\_end|>}. The dominant sinks are again all template tokens, so
the chat-template sink carries over to this single-stream model despite its
different encoder and architecture.

\begin{figure}[!htbp]
\centering
\includegraphics[width=\textwidth]{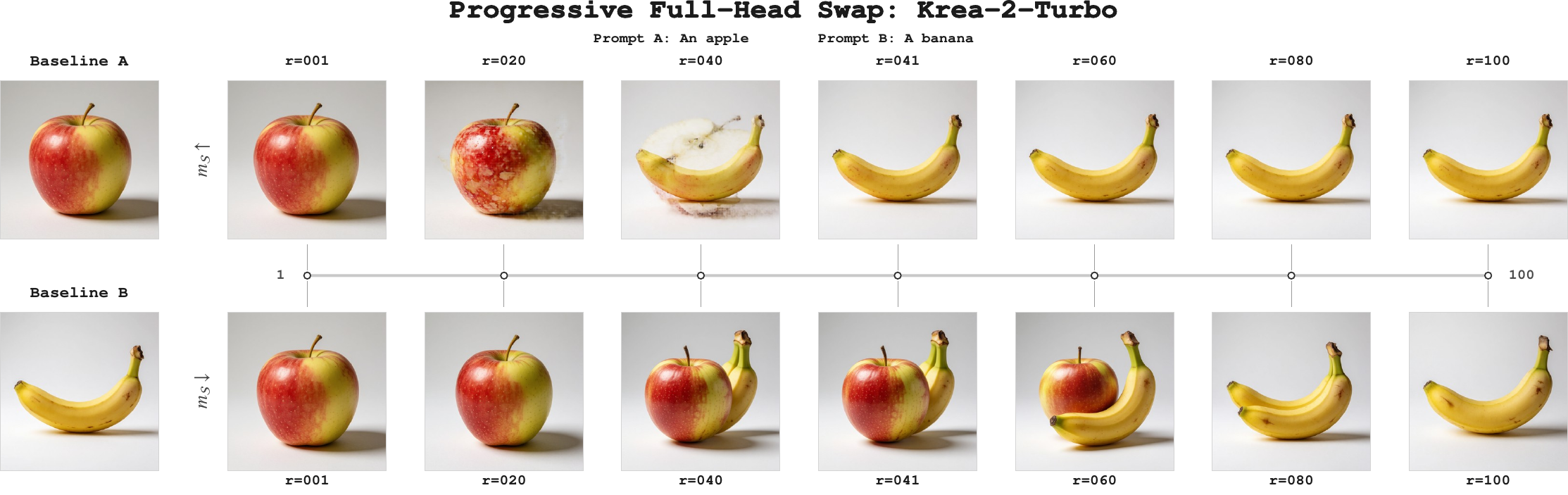}
\vspace{-0.8em}
\caption{Results for the progressive head-swap experiment on Krea-2-Turbo.}
\label{fig:krea2}
\end{figure}

Fig.~\ref{fig:krea2} reports the progressive head-swap experiment on
Krea-2-Turbo, whose outcome is consistent with our conclusion.

\subsection{Extending the Analysis to Distilled and Edit Models}

\paragraph{Few-step distillation.}
If the template span truly acts as a register, its attention signature should
be a stable part of the model's computation rather than an incidental feature of
the long multi-step trajectory. Few-step distillation is a natural stress test.
It collapses the $50$-step sampler into a $2$-step one and substantially alters
the denoising dynamics. On the same $2$-step distilled LoRA, now run with its
native $2$-step schedule, we recompute the I2T statistics of Sec.~\ref{sec:sinks} over GenEval. The sink survives essentially
intact. Template tokens still absorb $77\%$ of all I2T
attention (per-token ratio $6.8\times$), outweigh the entire semantic span at
$89\%$ of $(t,l,h)$ sites, and \texttt{<|im\_end|>} remains the single dominant
sink, with each statistic within a point of the undistilled checkpoint.
This indicates the register is intrinsic to the transformer's computation rather
than an artifact of any particular sampling schedule.

\paragraph{Instruction-based editing.}
A complementary stress test changes the conditioning regime rather than the
sampler, adding entirely new conditioning blocks. The editing model
Qwen-Image-Edit-2511 is conditioned on a reference image that enters
twice: in the text stream as a block of vision tokens $\mathcal{V}$ produced by
the VLM encoder, and in the image stream as VAE latents $z_{\mathrm{ref}}$
appended after the generated latents $z_{\mathrm{gen}}$. The full sequence
processed by joint attention thus becomes, in token order,
$[\,\mathcal{V};\mathcal{S};\mathcal{R};z_{\mathrm{gen}};z_{\mathrm{ref}}\,]$.
The template tokens now compete not only with the instruction
$\mathcal{S}$ but also with two representations of the reference image---the
vision tokens $\mathcal{V}$ and the latents $z_{\mathrm{ref}}$---that carry the
actual visual content of the edit. We recompute the I2C (Conditions) statistics over GEdit-Bench~\citep{liu2025step1x}. Per token,
$\mathcal{R}$ remains the single strongest attractor of the whole conditioning
sequence. Each template token draws $7.7\times$ the attention of a vision token
in $\mathcal{V}$, $13.8\times$ that of an instruction token, and $194\times$ that
of a reference-image latent in $z_{\mathrm{ref}}$, exceeding the per-token mass of
all three spans in $100\%$ of edits, with \texttt{<|im\_end|>} again the dominant
template sink. Thus, even when a new and highly informative conditioning
modality is added, the template span retains its register role per token,
reinforcing that the sink is intrinsic to the transformer's computation rather
than tied to any particular task.

\subsection{Extra Cases for Cross-Prompt Swaps}

\begin{figure}[!htbp]
\centering
\includegraphics[width=\textwidth]{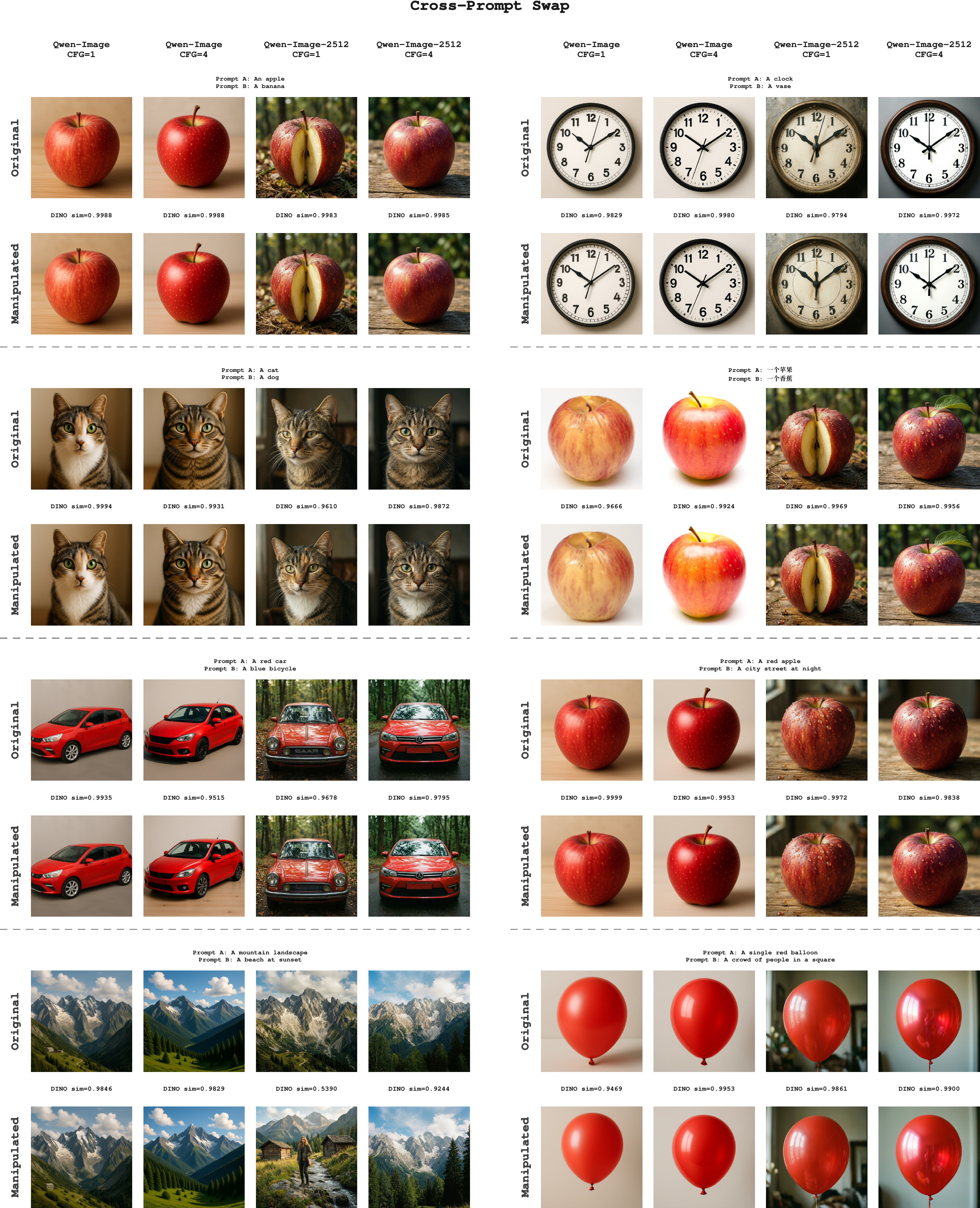}
\vspace{-0.8em}
\caption{Additional results for the cross-prompt swap experiment in Fig.~\hyperref[fig:swapmean]{\ref*{fig:swapmean}a}.}
\label{fig:extended_swap}
\end{figure}

Fig.~\ref{fig:extended_swap} reports additional prompt pairs for the
cross-prompt swap of Fig.~\hyperref[fig:swapmean]{\ref*{fig:swapmean}a}. For each
pair $(A,B)$ we build, at a fixed seed and sampler, the template-token swap
$[\mathcal{S}^{A};\mathcal{R}^{B}]$, which keeps $A$'s content tokens but
substitutes $B$'s template tokens. Across every pair the template-token swap
preserves the semantics of prompt $A$ and differs at most in
prompt-irrelevant content not specified by $A$.

\subsection{Extra Cases for Template-Token Averaging}

\begin{figure}[!htbp]
\centering
\includegraphics[width=\textwidth]{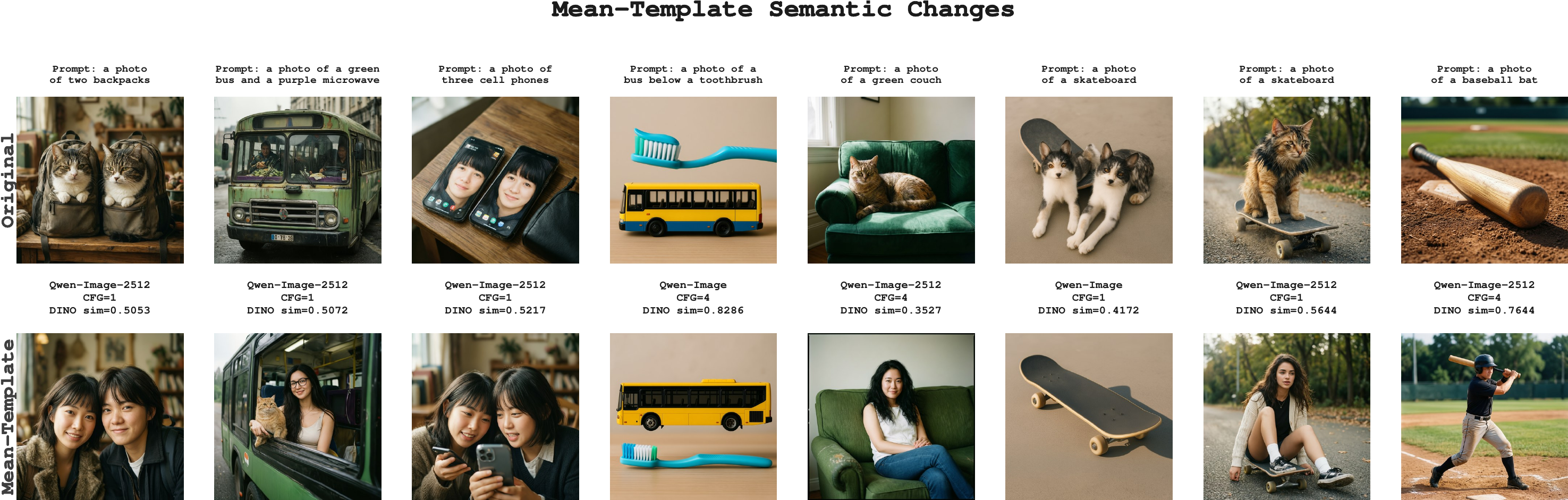}
\vspace{-0.8em}
\caption{Additional results for the template-token averaging experiment in Fig.~\hyperref[fig:swapmean]{\ref*{fig:swapmean}b}, showing cases where replacing $\mathcal{R}$ with the shared average $\bar{\mathcal{R}}$ induces large semantic changes.}
\label{fig:extended_mean}
\end{figure}

Fig.~\ref{fig:extended_mean} collects additional examples for the
template-token averaging experiment of
Fig.~\hyperref[fig:swapmean]{\ref*{fig:swapmean}b}, where each prompt's
template span is replaced by a single prompt-agnostic average $\bar{\mathcal{R}}$
and grafted onto the prompt's own semantics $[\mathcal{S};\bar{\mathcal{R}}]$. For
the vast majority of prompts the object is unchanged, so we focus here on the rare
tail of low-similarity cases in which the shared template induces large semantic
changes.

\subsection{Seed Sweep for Template-Token Averaging}

\begin{figure}[!htbp]
\centering
\includegraphics[width=0.75\textwidth]{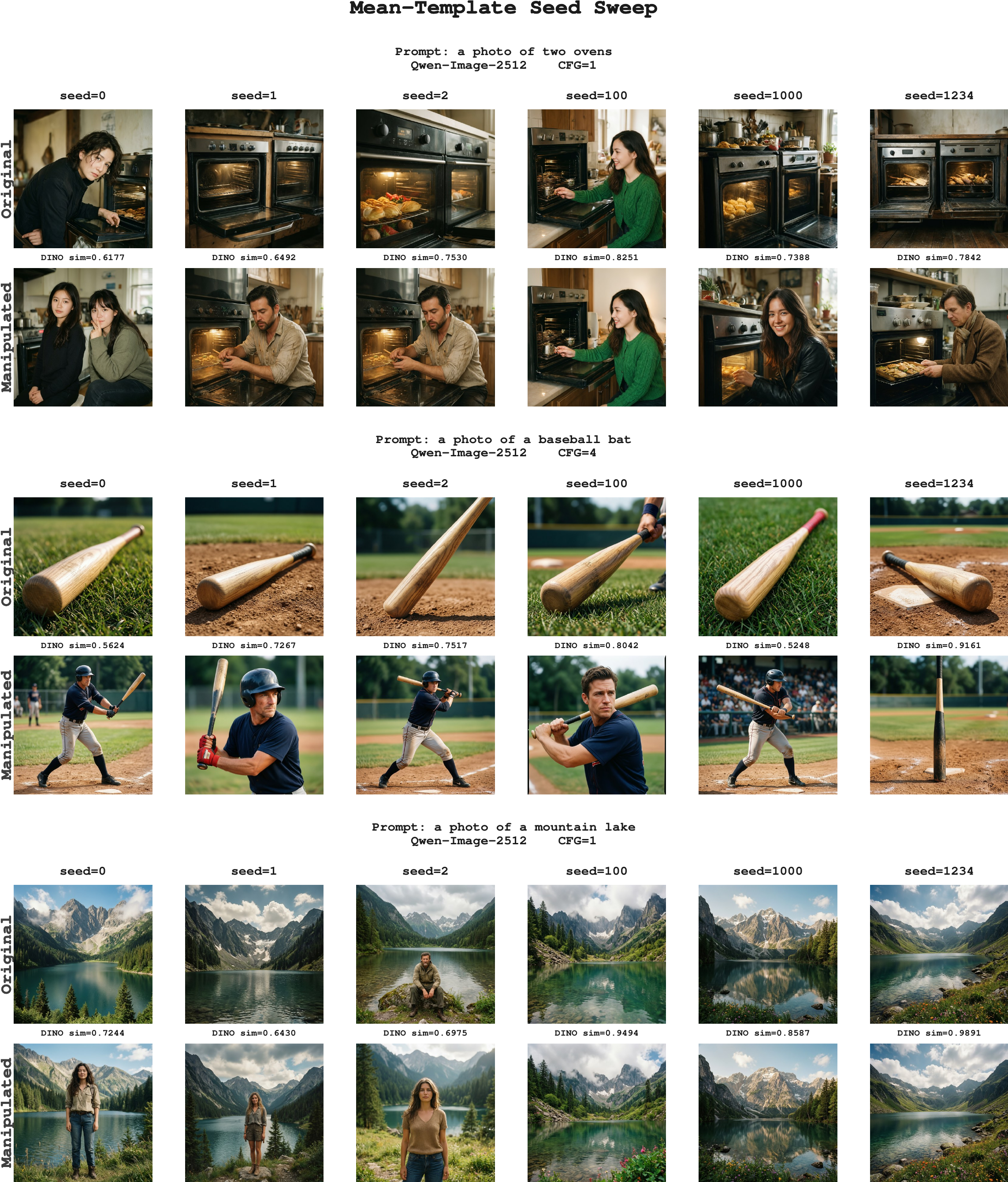}
\caption{Seed sweep for the template-token averaging experiment of Fig.~\hyperref[fig:swapmean]{\ref*{fig:swapmean}b}. The degree to which this averaging affects the semantics varies across prompts.}
\label{fig:seed_mean}
\end{figure}

Fig.~\ref{fig:seed_mean} sweeps the random seed for the template-token
averaging experiment of Fig.~\hyperref[fig:swapmean]{\ref*{fig:swapmean}b},
regenerating $[\mathcal{S};\bar{\mathcal{R}}]$ across seeds to separate the
semantic effect of the shared average $\bar{\mathcal{R}}$ from ordinary sampling
variation. The degree to which $\bar{\mathcal{R}}$ perturbs the semantics
varies markedly across prompts.

\subsection{Extra Cases for Progressive Head Swap}

\begin{figure}[!htbp]
\centering
\includegraphics[width=\textwidth]{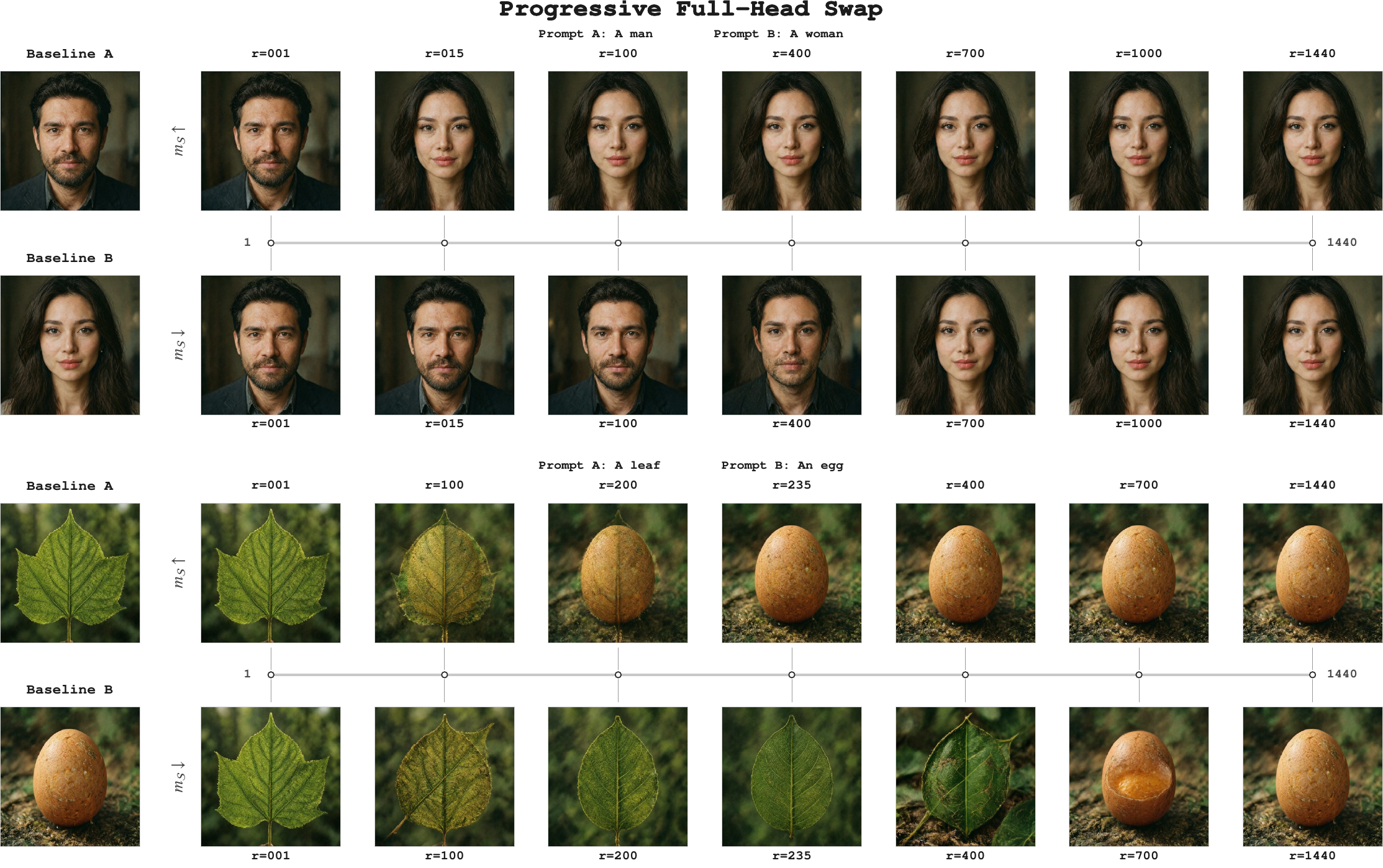}
\caption{Additional results for the progressive head-swap experiment in Fig.~\hyperref[fig:semantic]{\ref*{fig:semantic}a}, showing the same behavior across further prompt pairs.}
\label{fig:manwoman_leafegg}
\end{figure}

Fig.~\ref{fig:manwoman_leafegg} reports additional prompt pairs for the
progressive head-swap experiment of
Fig.~\hyperref[fig:semantic]{\ref*{fig:semantic}a}, where heads are swapped in
order of their semantic-span attention $m_\mathcal{S}$. Across every pair the
outcome is the same: swapping the top semantic readers first fails to transfer
the identity, while the reverse order flips the object after only a small
fraction of the heads, confirming that reading $\mathcal{S}$ and carrying the
object are decoupled.

\subsection{Per-Span Progressive Swap}

The progressive head-swap of
Fig.~\hyperref[fig:semantic]{\ref*{fig:semantic}a} transplants per-head
projections $(q,k,v)$ ranked by attention. To attribute the effect to a specific
conditioning span rather than to individual heads, we instead swap the full
$(q,k,v)$ of one entire span at a time from the $B{=}$``A banana'' run into the
$A{=}$``An apple'' run, progressively over the transformer blocks. As in the head-swap experiment, we run on the
$2$-step distilled LoRA at a single denoising step.

\begin{figure}[!htbp]
\centering
\includegraphics[width=\textwidth]{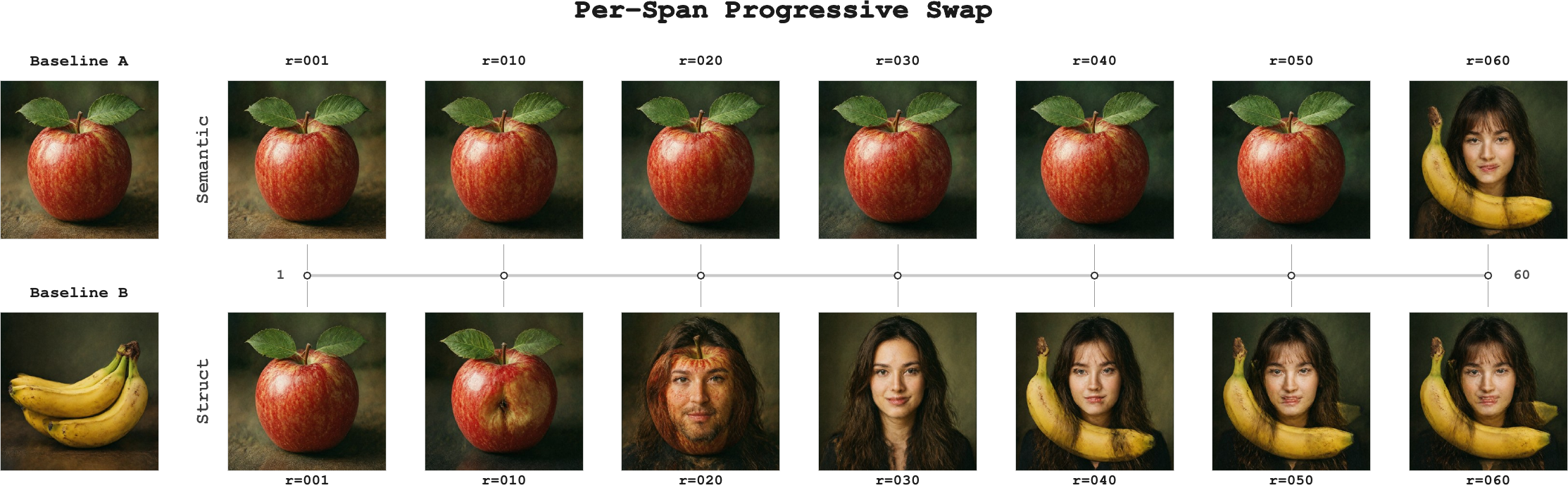}
\caption{Per-span progressive swap. The $(q,k,v)$ of one span---semantic
$\mathcal{S}$ (top) or template $\mathcal{R}$ (bottom)---is copied from $B$ into
$A$ over the first $r$ blocks. Swapping $\mathcal{S}$ preserves the apple; swapping
$\mathcal{R}$ disrupts it early.}
\label{fig:block_qkv_swap}
\end{figure}

As shown in Fig.~\ref{fig:block_qkv_swap}, swapping the template span
$\mathcal{R}$ changes the image drastically after only a fraction of the blocks,
whereas swapping the semantic span $\mathcal{S}$ leaves it almost unchanged until
nearly all blocks are swapped. In neither case is the target banana reproduced.
With the image tokens left untouched, a text-side swap alone, even at full
depth, only pushes the image toward the unconditional default, unlike the head
swap of Fig.~\hyperref[fig:semantic]{\ref*{fig:semantic}a}, which also transplants
the image-token projections. We can therefore conclude that part of the object's semantics is carried by the
semantic span $\mathcal{S}$ and, in particular, the image tokens, rather than by
the template register $\mathcal{R}$ alone.

\subsection{Qualitative Results of Head Pruning}

\begin{figure}[!htbp]
\centering
\includegraphics[width=\textwidth]{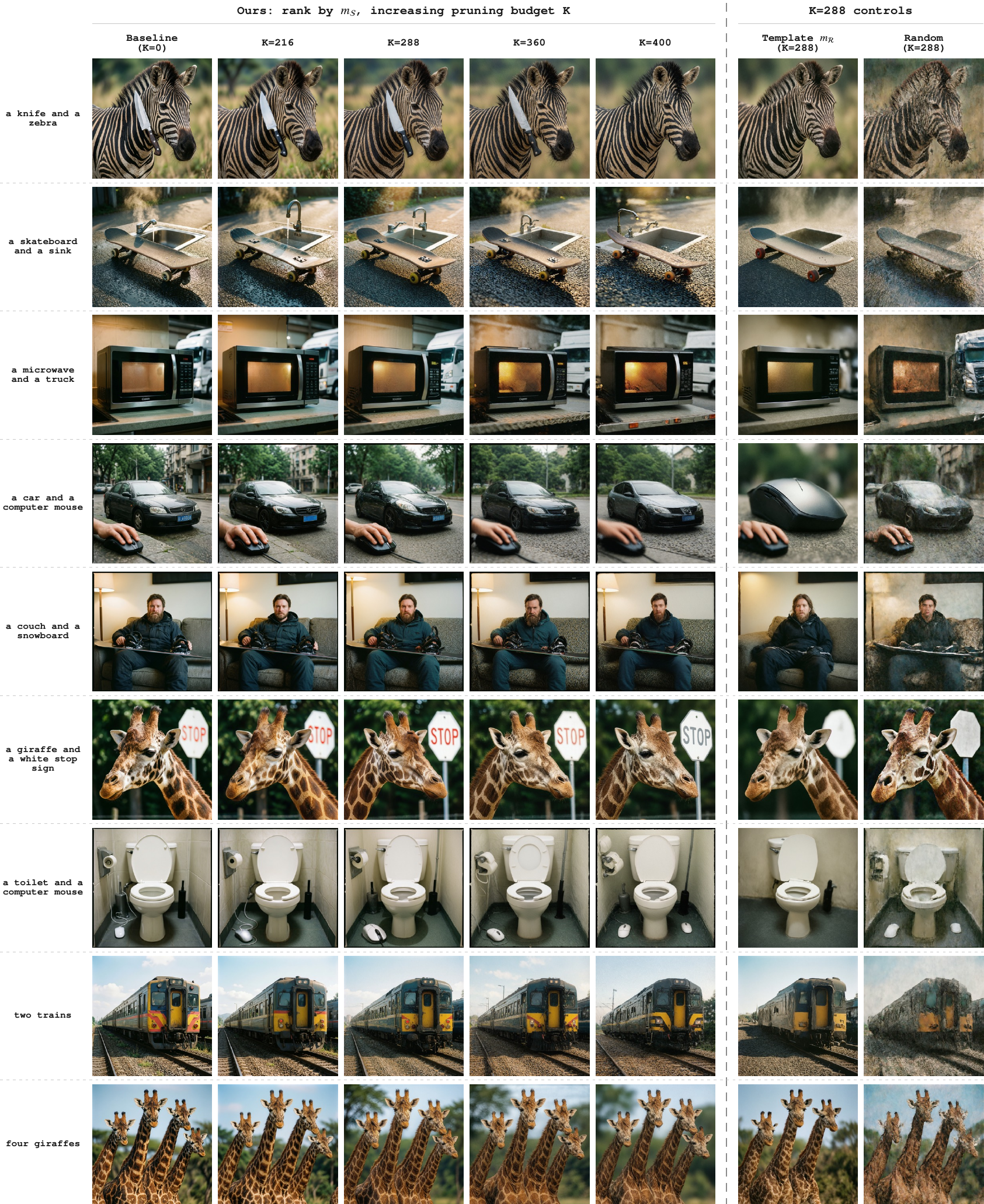}
\caption{Qualitative results of head pruning.}
\label{fig:pruning}
\end{figure}

Fig.~\ref{fig:pruning} shows generations at increasing pruning budgets $K$, illustrating how object identity is preserved while perceptual quality degrades gradually as more high-$m_{\mathcal S}$ heads are silenced.

\subsection{Distinct Mechanisms at Different Heads and Depths}

\begin{figure}[!htbp]
\centering
\includegraphics[width=0.75\textwidth]{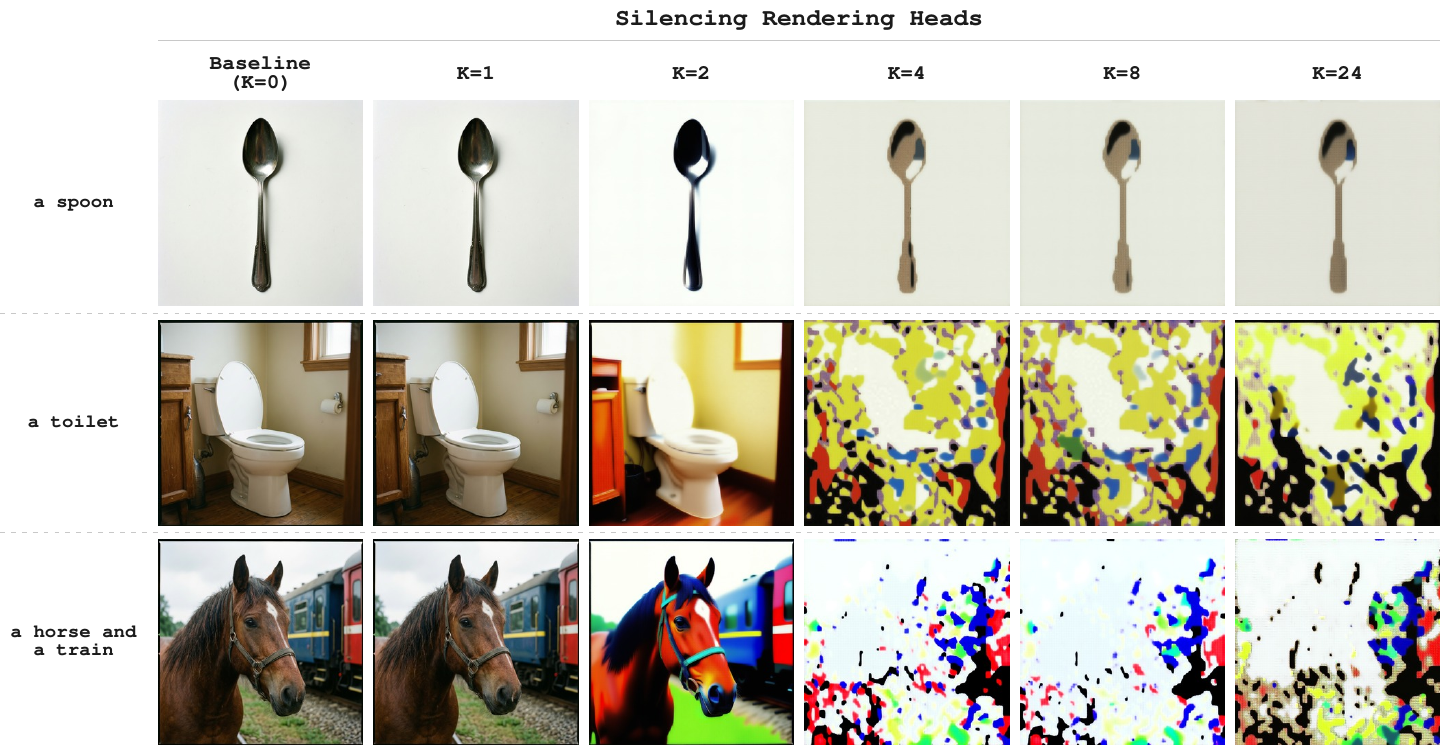}
\caption{Qualitative results of silencing rendering heads.}
\label{fig:render}
\end{figure}

Fig.~\ref{fig:render} provides the qualitative rendering-head ablation results
discussed in the main text. Silencing even a small fraction of heads whose
attention concentrates in the I2I block collapses the output into incoherent
patches, distinguishing their rendering role from the identity-carrying role of
semantic register heads.

\begin{figure}[!htbp]
\centering
\includegraphics[width=0.75\textwidth]{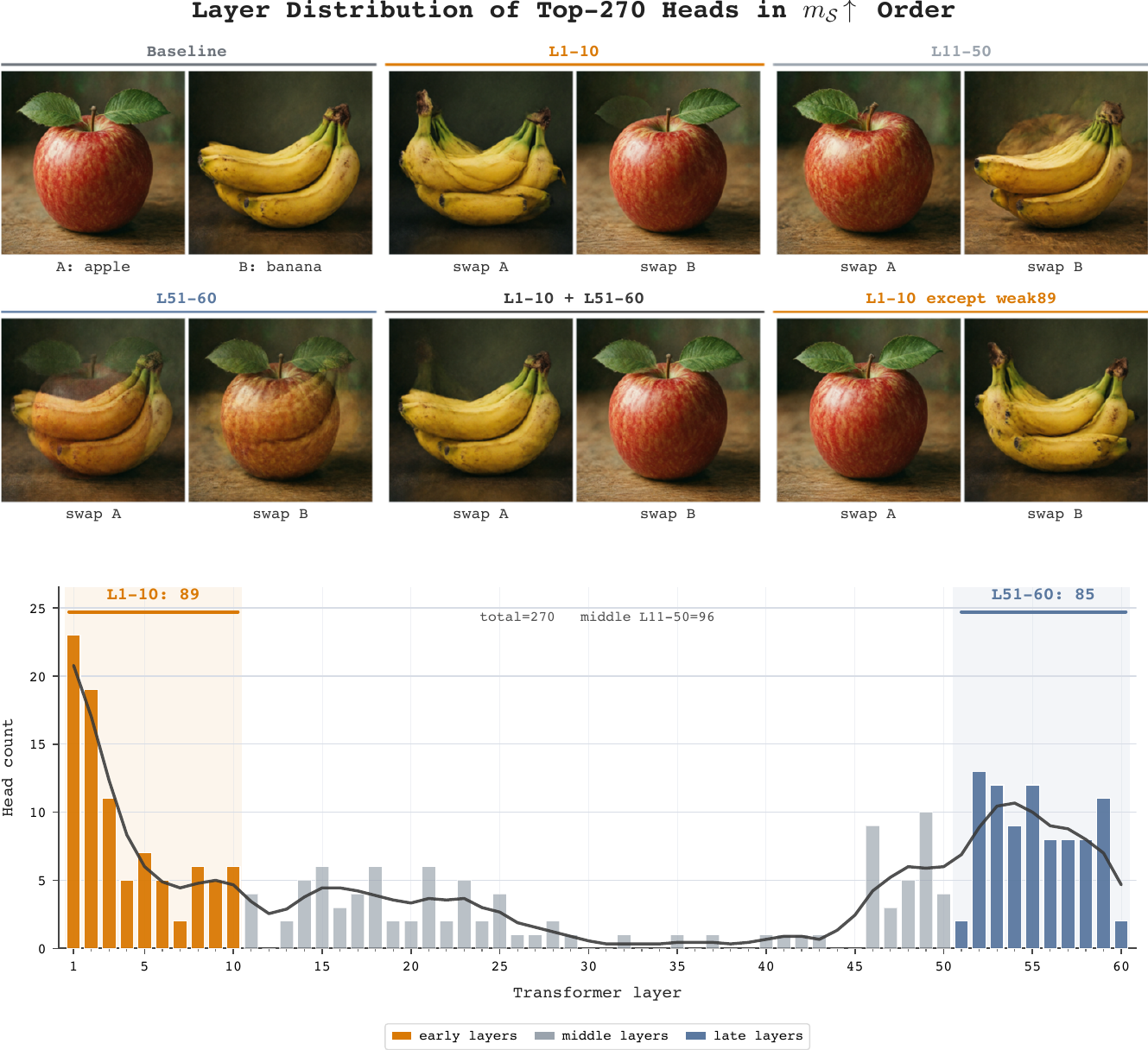}
\caption{Layer distribution of the first 270 heads in the reverse semantic-attention order ($m_{\mathcal S}\uparrow$). Top panels show the swap results for different layer groups; bottom shows how these heads are distributed across transformer layers. The heads concentrate in early layers and again in late layers, matching the layer groups that most strongly affect the apple-to-banana identity transfer.}
\label{fig:layer}
\end{figure}

Fig.~\ref{fig:layer} localizes the first $270$ heads in the reverse
semantic-attention order across depth. Their bimodal concentration in early and
late layers accompanies the early-commit, middle-carry, and late-refinement
division described in the main text.

\subsection{Temporal and Depth Structure of Attention Sinks}

\begin{figure}[!htbp]
\centering
\includegraphics[width=\textwidth]{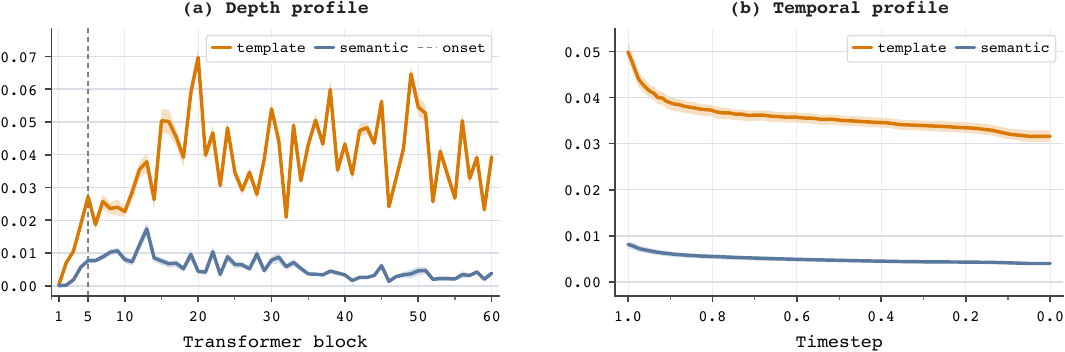}
\caption{Temporal and depth structure of the template sink on Qwen-Image-2512, aggregated over GenEval prompts.}
\label{fig:depth_temporal}
\end{figure}

Across GenEval prompts, template tokens receive much more I2T attention than
semantic tokens throughout denoising (Fig.~\ref{fig:depth_temporal}). On Qwen-Image,
template per-token mass falls from about $0.050$ at
$t{=}1.0$ to $0.035$ near $t{=}0$, while semantic-token mass falls from
$0.009$ to about $0.004$. The advantage therefore grows from
$5.3\times$ to $7.7\times$ over denoising.
The depth profile may also reveal the sink's origin. Template per-token mass
is near zero in the first block, reaches half its later-block plateau only at
block $11$, peaks around block $20$, and remains substantial later; onset
varies little across prompts. An input artifact such as large-norm template
embeddings should appear in the first block and remain roughly flat with depth.
Near-zero early blocks thus argue against an input-driven effect, while
prompt-independent onset would be unlikely if content-driven. Together, this
suggests a learned, depth-localized mechanism assembled by early-to-middle
blocks rather than inherited from the conditioning embeddings,
mirroring how attention sinks emerge in language models~\citep{gu2025attention}.

\end{document}